\documentclass[11pt,letterpaper,onecolumn]{IEEEtran}
\usepackage{setspace}
\usepackage{amsmath,amssymb,amsfonts}
\usepackage{mathtools}
\usepackage{dsfont}
\usepackage{array}
\usepackage{graphicx}
\usepackage{tikz}
\usepackage{subfigure}
\usepackage{color}
\usepackage{epsfig}
\usepackage{latexsym}
\usepackage{url}
\usepackage{cite}
\usepackage{comment}
\usepackage[margin=1in]{geometry}
\usepackage{float}
\usepackage{needspace}
\usepackage[section]{placeins}
\usepackage{fancyhdr}

\makeatletter
\def\fps@table{H}
\makeatother
\usepackage{algorithm}
\usepackage{algorithmic}
\newcommand{\algline}[1]{%
\STATE \parbox[t]{\dimexpr\linewidth-\algorithmicindent\relax}{\raggedright #1}%
}

\newcolumntype{L}{>{\raggedright\arraybackslash}m{1.7cm}}
\newcolumntype{M}{>{\raggedright\arraybackslash}m{1.7cm}}
\newcolumntype{N}{>{\raggedright\arraybackslash}m{3.808cm}}

\makeatletter
\g@addto@macro\normalsize{%
  \setlength\abovedisplayskip{4pt plus 1pt minus 2pt}%
  \setlength\belowdisplayskip{4pt plus 1pt minus 2pt}%
  \setlength\abovedisplayshortskip{2pt plus 1pt minus 1pt}%
  \setlength\belowdisplayshortskip{2pt plus 1pt minus 1pt}%
}
\makeatother

\newtheorem{theorem}{Theorem}%[section]

\newtheorem{example}{Example}
\newtheorem{assumption}{Assumption}%[section]

\newtheorem{lemma}{Lemma}
\newtheorem{definition}{Definition}

\newtheorem{corollary}{Corollary}

\newcommand{\fsquare}{\vrule height6pt width7pt depth1pt}

\newcommand{\myendpf}{\hfill\fsquare \\[0.1in]}

\allowdisplaybreaks

\newcommand{\pr}{\mathbb{P}}
\newcommand{\E}{\mathbb{E}}

\newcommand{\R}{\mathbb{R}}

\newcommand{\N}{n}
\newcommand{\T}{t}
\newcommand{\pz}{p_z}
\newcommand{\wprime}{w}

\newcommand{\zprime}{{+1}}
\newcommand{\zdprime}{{-1}}

\newcommand{\es}{s}
\newcommand{\esprime}{s'}
\newcommand{\Yk}{{Y}^{(k)}}

\newcommand{\yk}{{y}^{(k)}}
\newcommand{\Zk}{{Z}^{(k)}}
\newcommand{\zk}{{z}^{(k)}}
 \newcommand{\gammak}{{\gamma}^{(k)}}

\newcommand{\Thetahat}{\widehat{\Theta}}

\newcommand{\gammakhat}{\widehat{\gamma}^{(k)}}

\newcommand{\delsep}{\Delta_{ \rm sep}}
\newcommand{\KL}[2]{D_{\mathrm{KL}}\!\left(#1 \,\|\, #2\right)}

\newcommand{\iprimehat}{\widehat{I}_{\zprime}}
\newcommand{\idprimehat}{\widehat{I}_{\zdprime}}

\makeatletter
\g@addto@macro\normalsize{%
  \setlength\abovedisplayskip{4pt plus 1pt minus 2pt}%
  \setlength\belowdisplayskip{4pt plus 1pt minus 2pt}%
  \setlength\abovedisplayshortskip{2pt plus 1pt minus 1pt}%
  \setlength\belowdisplayshortskip{2pt plus 1pt minus 1pt}%
}
\makeatother

\usepackage[caption=false,font=footnotesize]{subfig}
\newcommand{\losssingle}{\mathcal{L}_{\N,\T}\left(\{\yk\}_{k\in[\N]};\Theta,\{\gamma^{(k)}\}_{k\in[\N]}\right)}
\newcommand{\losssingleunif}{\mathcal{L}^\prime_{\widehat I_c,\T}\left(\{\yk\}_{k\in[\widehat I_c]};\Theta,\gamma\right)}
\newcommand{\lossidealized}{\widetilde{\mathcal{L}}\left(\Theta,\{\gammak\}_{k\in[\N]}\right)}
\DeclareMathOperator*{\argmin}{arg\,min}

\newcommand{\covernum}[3]{\mathcal{C}_{#1}\left(#3\right)}

\newcommand{\mentropy}[3]{\mathcal{M}_{#1}\left(#3\right)}

\newcommand{\Bn}[4]{\mathcal{B}^{{#1},\|.\|_{#2}}(#3,#4)}

\usepackage{xcolor}
\definecolor{mygray}{gray}{0.7}

\usepackage{tabularx}

\newcommand{\setz}{\mathbb{S}_Z}
\newcommand{\const}{\mathbb{C}}

\fancypagestyle{plain}{%
}
\begin{document}

\title{Learning Mixtures of Exponential Family}

\author{Mansi~Sood, Devavrat~Shah\\
Laboratory for Information \& Decision Systems\\
Massachusetts Institute of Technology%
\thanks{M. Sood and D. Shah are with the Laboratory for Information \& Decision Systems, Massachusetts Institute of Technology, Cambridge, MA, USA (e-mail: msood@mit.edu; devavrat@mit.edu). This work was supported in part by the Schmidt Science Fellowship and ONR MURI Award (N00014-24-1-2687).}
\thanks{AI Statement: The OpenAI model GPT-6-Astra available through Codex was used to verify the correctness of the proof. This led to minor
 corrections in few places. The entire manuscript is written by authors. The authors take full responsibility for {entirety} of it.}
} % Close \author here, after BOTH \thanks commands

\maketitle

% \author{Mansi~Sood, Devavrat~Shah\\  Laboratory for Information \& Decision Systems \\ Massachusetts Institute of Technology%\\
% \thanks{M. Sood and D. Shah are with the Laboratory for Information \& Decision Systems, Massachusetts Institute of Technology, Cambridge, MA, USA (e-mail: msood@mit.edu; devavrat@mit.edu). This work was supported in part by the Schmidt Science Fellowship and 
% ONR MURI Award (N00014-24-1-2687).}}
% \thanks{AI Statement: The OpenAI model GPT-6-Astra available through Codex was used to verify the correctness of the proof. This led to minor
% corrections in few places. The entire manuscript is written by authors. The authors take full responsibility for entirety of it.}

% \maketitle

\begin{abstract}
Motivated by causal inference and physical computing systems, we consider
the problem of learning mixture of two bounded-support exponential-family distributions
with a shared interaction parameter matrix and mixture {component-specific} 
linear parameter vectors. The unknown sequence of component labels belongs to a known structured
class. We propose an estimator based on a convex loss that avoids the
log-partition function; for each fixed label assignment, estimation is a
convex optimization problem. We show that the conditional population loss
uniquely identifies the interaction matrix and the linear parameter attached
to each sample. For a global empirical minimizer, we establish finite-sample
bounds on the interaction error in Frobenius norm and the average squared
linear-parameter error. The bounds depend on the logarithm of the number
of admissible label assignments, and, for fixed dimension and model bounds, imply consistency when this complexity
grows sublinearly with sample size. Additional occupancy and separation
conditions yield component and label recovery. 
\end{abstract}


% !TeX root = ../../main-v2.tex
\section{Introduction}

\begin{figure}[t]
    \centering
    \includegraphics[width=1.0\linewidth]
    {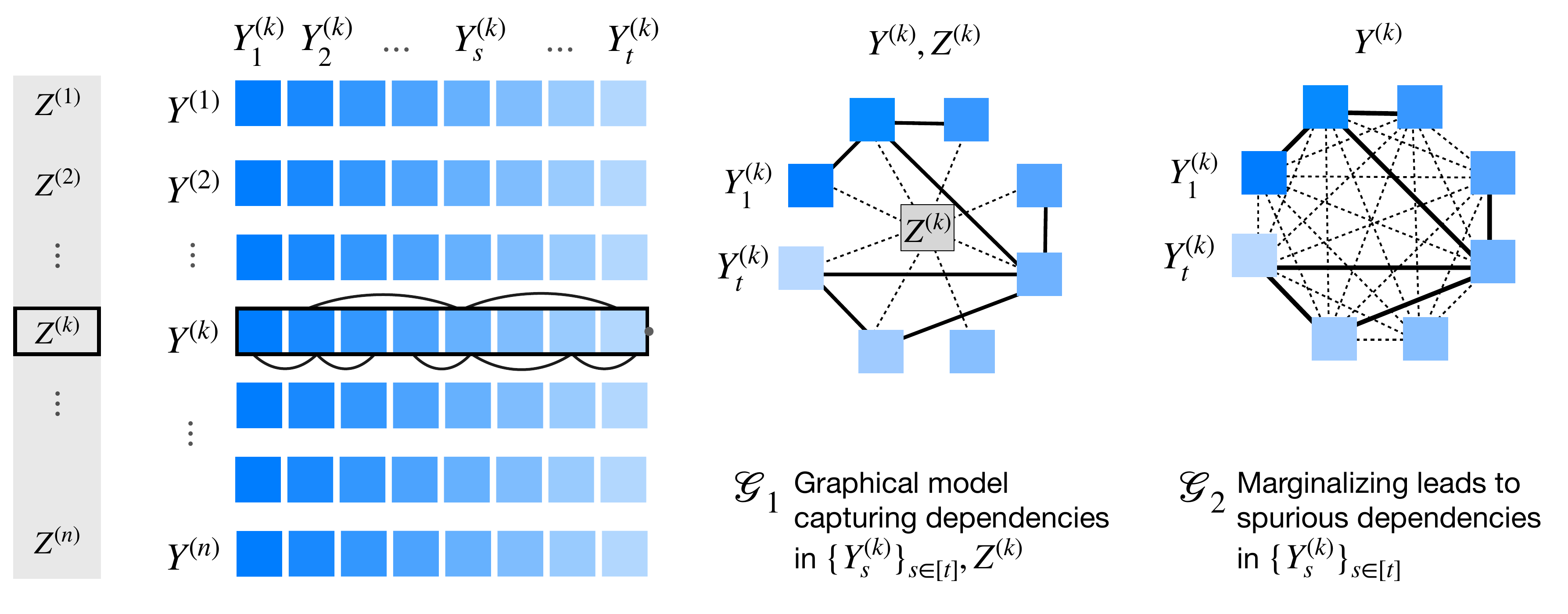}
    \caption{\footnotesize\setstretch{1.3}
    Mixture learning setup: Each sample $k\in[\N]$ comprises the pair
    $(Z^{(k)},Y^{(k)})$, where $Z^{(k)}$ is unobserved and $Y^{(k)}$
    is observed. In our model the label sequence is fixed or random within
    a known set $\setz(\N)$. Conditional on the full label sequence, the
    observations are independent, with the law of $Y^{(k)}$ determined by
    $Z^{(k)}$ and unknown exponential-family parameters. Marginalizing out $Z^{(k)}$
    can induce additional dependencies among the observed variables
    $\{Y_s^{(k)}\}_{s\in[\T]}$, resulting in a denser graphical model
    and complicating learning from the observed distribution.}
    \label{fig:model}
\end{figure}

\noindent The problem of learning mixture distributions with \emph{latent} types %structure 
serves as a cornerstone of inference in both natural and infrastructure systems
\cite{EM-royal,linday-IMS}. 
Dating back to Pearson's analysis of heterogeneous crab measurements
\cite{pearson1894iii,EM-royal}, mixture models provide a natural statistical
language for populations with {latent} subtypes or states. Failing to account
for such heterogeneity can lead to misleading inference, as classically
illustrated by Simpson's paradox \cite{simpson1951}, where relationships in an
aggregated population can differ from, or even reverse, those within the
underlying subpopulations. If the latent types were observed, learning could
proceed separately within each component. In observational settings, however,
only the outcomes are observed, and marginalizing over the latent types gives
rise to additional dependencies among the observed variables; see {\textbf{Figure~\ref{fig:model}}}.

Despite the ubiquity of such mixtures, learning them reliably and efficiently
from limited samples remains an important methodological challenge, requiring one to
simultaneously uncover the latent population structure and learn the
corresponding component distributions \cite{EM-royal,linday-IMS}. Progress in
the field has hinged on exploiting specific structure in the component family
to make statistically and computationally tractable recovery possible
\cite{dasgupta1999learning,linday-IMS}. 

We consider the framework where the distribution of the observed variables 
conditioned on the latent type belongs to the \emph{exponential family} with 
local and pairwise interactions.
The exponential-family framework encompasses many familiar discrete and
continuous distributions and provides a maximum-entropy interpretation
through constraints on sufficient statistics
\cite{wainwright2008graphical,barndorff2014information}. Our results concern
a specific subclass, with bounded product support, linear sufficient statistics,
and pairwise interactions with no diagonal quadratic terms. This includes
Ising models and bounded continuous pairwise models; it does not include
all exponential-family distributions without further modification.
Efficient learning within the exponential family has seen significant advances recently through
computationally tractable M-estimators, cf. \cite{Vuffray-lokhov,vuffray2020,shah2023computationally,
ren2022learningcontinuousexponentialfamilies}. Mixtures of exponential-family distributions, 
however, do not in general remain in the same exponential family, precluding 
the direct applicability of these methods to mixture learning.

\begin{figure}[htb]
    \centering
    {\setlength{\fboxsep}{0pt}%
     \setlength{\fboxrule}{0.3pt}%
     \fcolorbox{black!20}{white}{%
        \includegraphics[width=\dimexpr0.999\linewidth-2\fboxrule\relax]
        {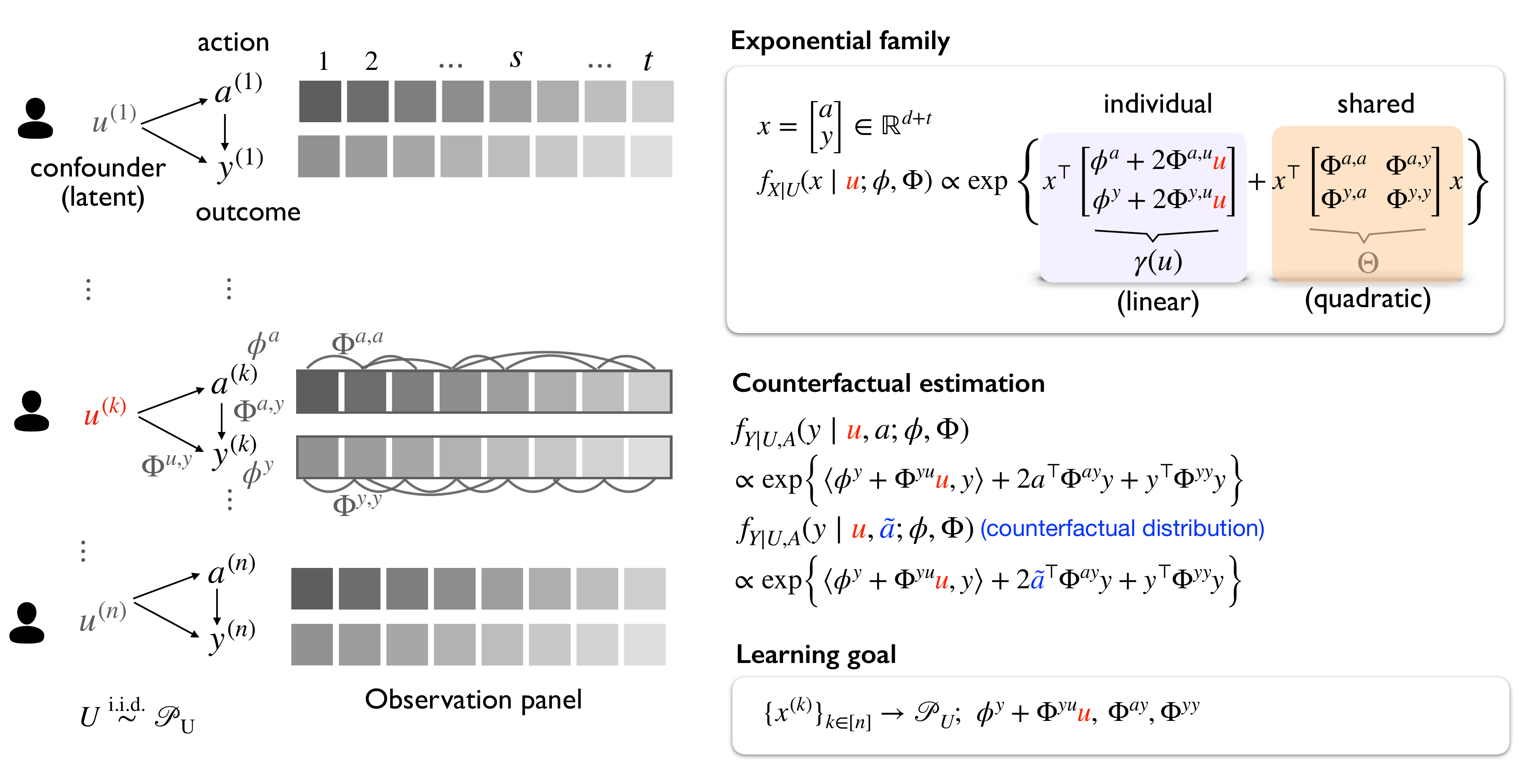}%
     }}
    \caption{\footnotesize\setstretch{1.3}
     \textbf{Causal Inference}:
    Independent units generate treatment-outcome panels from
    exponential-family: latent unit-level confounders $u^{(k)}$
    perturb the linear parameter $\gamma(u^{(k)})$, while the quadratic
    interaction parameter $\Theta$ remains shared across units.
    Learning the parameters for conditional distribution
    $f_{Y\mid U,A}$ supports counterfactual evaluation under alternative
    actions $\widetilde a$\cite{sood2026causal}.}
    \label{fig:motivate.causal}
\end{figure}

\begin{figure}[htb]
    \centering
    {\setlength{\fboxsep}{0pt}%
     \setlength{\fboxrule}{0.3pt}%
     \fcolorbox{black!20}{white}{%
        \includegraphics[width=\dimexpr0.999\linewidth-2\fboxrule\relax]
        {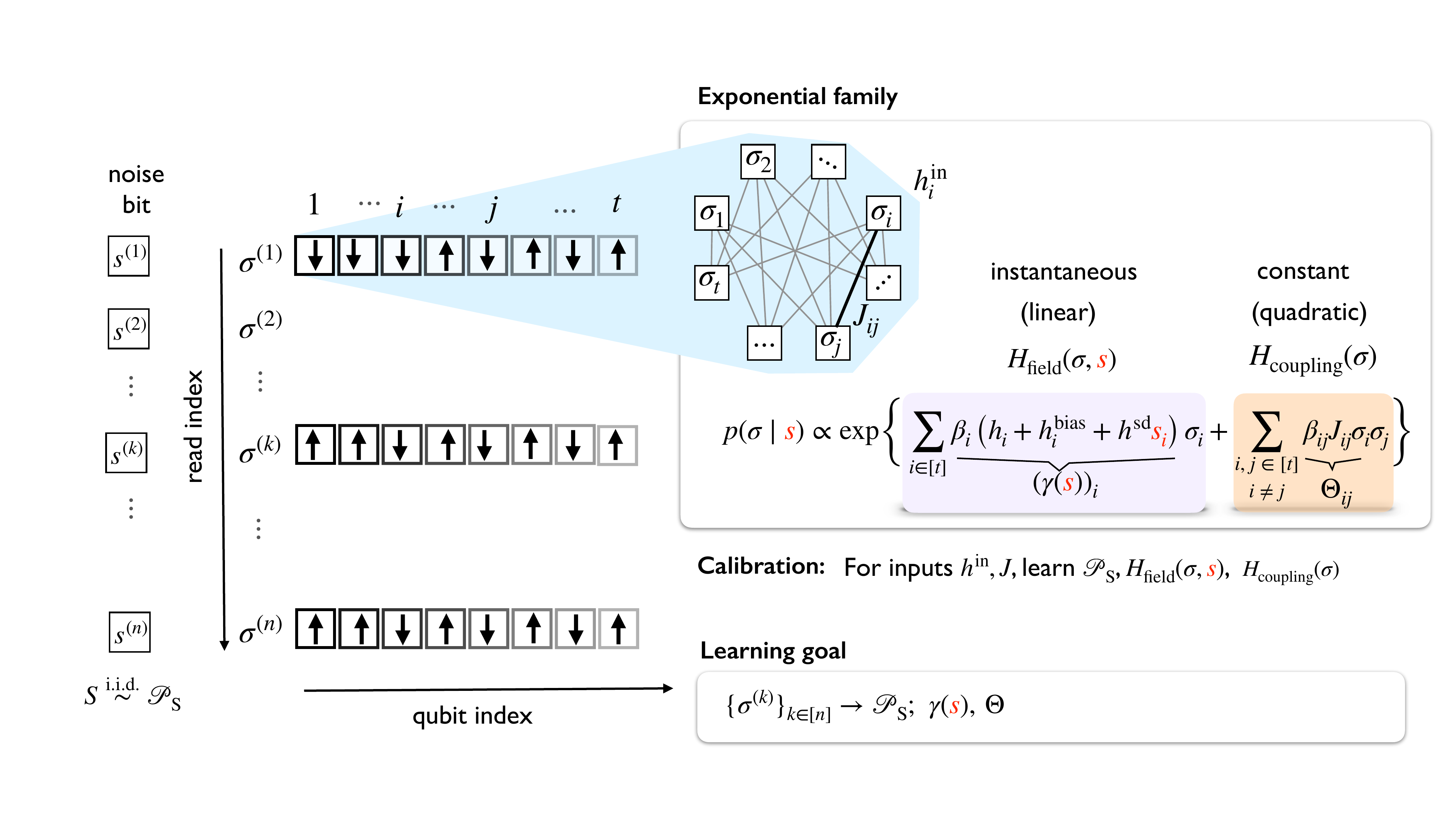}%
     }}
    \caption{\footnotesize\setstretch{1.3}
    \textbf{Quantum Computing}:
    Repeated anneals generate independent readouts  $\sigma^{(k)}$ from low-temperature
    noisy Gibbs distributions: Latent read-to-read noise $s^{(k)}$ perturbs the
    local fields $\gamma(s^{(k)})$, while the pairwise interaction
    parameter $\Theta$ encoding qubit interaction remains shared, as modeled in
    Fig.~\ref{fig:model-distributions}. Jointly learning the field and
    coupling Hamiltonians for given inputs $h^{\rm in},J$ by learning the
    parameters of exponential family $p(\sigma~|~s)$ is key to device calibration
    in quantum annealing\cite{PRXQuantum.3.020317}.}
    \label{fig:motivate.quantum}
\end{figure}

To facilitate learning, we further restrict attention to scenarios where \emph{pairwise}
interactions are \emph{shared} while \emph{local biases vary} across distributions 
conditioned on latent type.  This shared-interaction, varying-bias structure arises 
naturally in several applications.%; see Fig.~\ref{fig:motivate.causal}-\ref{fig:motivate.quantum}. 
In the setting of causal inference from observational data with panel of 
treatment-outcome, latent unit-level confounders induce variation in the linear parameters while the interaction structure 
remains shared across units, cf. \cite{shah2023counterfactualinferenceunobservedconfounding,sood2026causal}; see {\textbf{Figure~\ref{fig:motivate.causal}}}. 
In analog computing systems such as quantum annealers, read-to-read noise perturbs
local fields while pairwise couplings remain shared across anneals, cf. \cite{PRXQuantum.3.020317}; see {\textbf{Figure~\ref{fig:motivate.quantum}}}. 
Related forms of heterogeneity arise more broadly in interacting biological and social systems
\cite{isingmix-to-social,sood2026causal}. More formally, viewing the coordinates of each 
observation $Y^{(k)}\in\mathbb R^{\T}$ as nodes of a graphical model, the linear parameter 
$\gamma(Z^{(k)})$ captures local biases and varies with the latent type $Z^{(k)}$, while 
the quadratic interaction parameter $\Theta$ captures pairwise interactions and remains shared across
samples; see {\textbf{Figure~\ref{fig:model-distributions}}}.  For exponential-family models, this distinction 
is naturally represented through linear and quadratic sufficient statistics, corresponding 
respectively to local biases and pairwise interactions \cite{wainwright2008graphical}. 
We therefore focus on two component mixture distributions with linear and quadratic
sufficient statistics, and exploit a known class of admissible label
sequences whose logarithmic cardinality grows sublinearly with sample size. 

\medskip
\noindent \textbf{Contributions.}
We establish a joint estimation guarantee for a shared interaction parameter matrix
and component specific two linear parameter vectors, with unknown labels coming 
{from a} known class. Samples from both components contribute to estimation
of the interaction matrix, while observations assigned to the same
component share a linear parameter. The estimator minimizes a convex loss,
based on \cite{abhin-thesis}, that avoids the log-partition function.
Its feasible set is a union of convex sets indexed by the admissible
assignments. Thus a fixed assignment gives a convex problem, while global
computation depends on the size and structure of the known class to which
unknown labels belong.

Theorem~\ref{thm:idealized-loss} proves that the population loss conditional
on the true label sequence uniquely identifies the interaction matrix and
the sample-specific linear parameters. This is a conditional population
identification statement, rather than identification from an unrestricted
marginal mixture distribution. Theorem~\ref{thm:two-valued-linear-param}
then bounds the Frobenius interaction error and the average squared
linear-parameter error for any global empirical minimizer. The {sample} complexity is is that of one matrix and two vectors, plus logarithm of the size of the known class to which unknown labels belong.

% governed by estimating one shared interaction matrix and two component-specific linear parameter vectors, with an additional term given by the logarithm of the number of admissible label sequences.}
%The complexity
% is that of one matrix and two vectors, plus logarithm of the size of the known class
% to which unknown labels belong.
For fixed dimension and parameter bounds, this yields
consistency. 

When both components are represented and the error is sufficiently small
relative to their separation and empirical proportions,
Corollary~\ref{cor:pooled-components} gives component estimates and a bound
on the fraction of incorrect labels, up to permutation. Exact label
recovery requires an additional condition on the separation of admissible
partitions. 

We provide an illustrative example of results in Example~\ref{ex:change-point}. Here
we consider setting where observations come from first mixture component up to some number of
samples and then come from the second mixture component. Therefore, in effect 
it boils down to identifying this {\em change point} and subsequently the parameters 
associated with both the components. For such an example, our result provides a computationally tractable algorithm that has estimation error decreasing to arbitrary accuracy as long as number
of observations scale polynomially in the dimension of the observations. 

\medskip
\noindent \textbf{Related Work.}
Our work builds on literature on graphical models and exponential families
\cite{wainwright2008graphical,barndorff2014information}, computationally
efficient M-estimators for high-dimensional exponential-family models
\cite{Vuffray-lokhov,vuffray2020,shah2023computationally,
ren2022learningcontinuousexponentialfamilies,abhin-thesis},
and learning Ising models
\cite{dagan2020learningisingmodelsmultiple,kandiros21a,bresler2015efficiently,
klivans2017learning,tandonravikumar2014information}.
With few exceptions, much of this literature considers learning a single exponential-family
distribution; in contrast, this work considers observations generated from a latent
mixture while exploiting an interaction parameter shared across mixture
components.

More broadly, approaches to mixture learning fall into a few canonical
viewpoints. A per-sample empirical estimator takes each sample's own
maximum-likelihood estimate as a stand-in for its latent parameter and
approximates the mixing distribution by the empirical CDF of these estimates
\cite{wasserman-book}; %it is computationally trivial, but 
accuracy of such an approach is bottlenecked by within-sample noise. 
Moment-based estimators \cite{tian2017learning}
instead recover the mixing distribution by matching empirical moments to a
probability measure over the parameter space, typically via a linear program,
sidestepping likelihood evaluation altogether. Non-parametric maximum likelihood
estimation (NPMLE) \cite{icml-vinayak19a,lindsay1983geometry} discretizes the
latent support into bins to pool information across samples, performing a
reverse-I-projection of the observed sufficient-statistic fingerprint onto the
convex hull of the component family, though, like full likelihood-based methods,
it still requires evaluating normalizing constants. The EM algorithm
\cite{EM-royal,linday-IMS,balakrishnan2017statistical} instead retains the full
likelihood and alternates between computing expected latent assignments (E-step)
and updating parameters to maximize the expected complete-data log-likelihood
(M-step), iteratively constructing and maximizing a lower bound
of the log-likelihood; it remains sensitive to initialization owing to the
non-convex likelihood landscape. For Gaussian mixtures specifically, a related
two-step approach first projects the data into a low-dimensional subspace and
then clusters within it \cite{dasgupta1999learning}. Mixture learning also
admits a matrix-completion perspective {where samples} are stacked into an observation
matrix whose conditional mean has one distinct row per mixture component
inducing a low-rank, clustered-row structure, which spectral methods can exploit
to recover latent parameters directly, bypassing both likelihood maximization
and moment matching
\cite{AgarwalShahShen2020,BorgsChayesShahYu2017,ChenChiFanMa2020}.

Our estimator replaces likelihood evaluation with a convex loss that
sidesteps the log-partition function. It pools parameters within each
candidate assignment and selects a global minimizer over the admissible
assignment class. 
The two-valued { component} structure is enforced during estimation;
no separate clustering stage is needed. 
% The estimator jointly estimates the two component-specific linear parameter vectors and assigns samples to components, so no separate clustering stage is needed. 
The analysis exploits this
structure through a covering argument for one matrix and two vectors,
followed by a union bound over assignments. General {i.i.d.} two-component
labels can admit $2^{\N}$ possible sequences and do not satisfy assumptions 
required for our result. The present guarantees therefore concern
structured latent heterogeneity, motivated in part by the shared-parameter
models arising in causal inference
\cite{sood2026causal,agarwal2025synthetic,shah2023counterfactualinferenceunobservedconfounding}.

% A key computational feature of our approach is that estimation is performed
% through a convex objective that avoids direct optimization of the mixture
% likelihood and computation of the log-partition function. This provides an
% alternative to likelihood-based procedures such as the expectation-maximization
% (EM) algorithm \cite{balakrishnan2017statistical}, whose optimization
% landscape is generally non-convex. More generally, our method can interpolate
% between regimes with rich prior knowledge of the parameter space but limited
% access to samples, and regimes with multiple independent samples but limited
% assumptions on the parameter space
% \cite{dagan2020learningisingmodelsmultiple,kandiros21a}.\\

\medskip
\noindent \textbf{Organization.}
Section~\ref{sec:setup} introduces the model, the learning problem of interest,
and the key operating assumptions. Section~\ref{sec:results} presents the
estimation algorithm and its performance guarantees, including identifiability
(Theorem~\ref{thm:idealized-loss}), joint recovery of the interaction matrix
and the sample-averaged linear parameters
(Theorem~\ref{thm:two-valued-linear-param}), and component
estimation and label recovery (Corollary~\ref{cor:pooled-components}).
Example~\ref{ex:change-point} specializes the bound to an unknown change point. All proofs are presented in the
appendix.

\medskip
\noindent\textbf{Notation}. For a natural number $\N$, define $[\N] := \{1,2,\ldots,\N\}$.
We write $\{\square^{(k)}\}_{k \in [\N]}$ for the collection
$\{\square^{(1)}, \cdots, \square^{(\N)}\}$ indexed by $k \in [\N]$.
We use the superscript ${\square}^{(k)}$ to index over samples and the subscript
${{\square}^{(k)}}_{t}$ to denote the $t^{\text{th}}$ coordinate of the $k^{\rm th}$ sample.
For $\Lambda_\gamma \subseteq \mathbb{R}^{\T},  $ let
$\Lambda_\gamma^{[\N]}:= \Lambda_\gamma\times\dots \times \Lambda_\gamma $
($\N$ times).  For a vector ${v} \in \mathbb{R}^{\T}$, we use $v_\es$ to denote
its $\es^{\text{th}}$ coordinate and ${v}_{-\es} \in \mathbb{R}^{\T-1}$ to denote
the vector obtained after deleting the $\es^{\text{th}}$ coordinate. We denote the
$\ell_0$, $\ell_p$ $(p \ge 1)$, and $\ell_\infty$ norms of a vector ${v}$ by
$\|{v}\|_0$, $\|{v}\|_p$, and $\|{v}\|_\infty$, respectively. %For $u,v \in \R^\T$, 
For $u,v \in \R^\T$, ${\rm{MSE}}(u,v)= \T^{-1} \sum_{\es \in [\T]} (u_\es-v_\es)^2$.
For a matrix ${M} \in \mathbb{R}^{\N \times \T}$, we denote the $\es^{\rm th}$
row as ${M}_\es$ and the element corresponding to the $\es^{\rm th}$ row and
$\esprime^{\rm th}$ column is ${M}_{\es \esprime}$. Further, denote
$\esprime^{\rm th}$ column as $M(:,\esprime)$. 
%We denote the matrix maximum norm by $\|{M}\|_{\max}$, the induced
%$\infty$-norm (operator $\infty$-norm) by $\|{M}\|_{\infty}$, and the
%$(2,\infty)$-norm by $\|{M}\|_{2,\infty}$.
We denote the matrix maximum norm by $\|M\|_{\max}$,
the induced $\infty$-norm (operator $\infty$-norm) by $\|M\|_{\infty}$,
and the $(2,\infty)$-norm by $\|M\|_{2,\infty}$,
{ where $\|M\|_{2,\infty}:=\max_i\|M_i\|_2$.
We also denote $\|M\|_{1,1}:=\sum_{i,j}|M_{ij}|$.}
We take $\mathcal Y$ to be either a finite subset of $\mathbb R$ with at
least two points, or a bounded measurable subset with positive Lebesgue
measure. Let $\mathcal U_{\mathcal Y}$ be its uniform probability measure
(normalized counting measure in the finite case, normalized Lebesgue
measure otherwise), and write
$\mu_{\mathcal U}:=\mathbb E_{Y\sim\mathcal U_{\mathcal Y}}[Y]$.
All densities on $\mathcal Y^d$ are relative to
$\mathcal U_{\mathcal Y}^{\otimes d}$. Integrals written with $dy$ use this
product reference measure; in the finite case they are finite sums with
uniform weights. Thus the same notation covers discrete and continuous
observations.
For probability measures $P$ and $Q$, s.t., $P$ is absolutely continuous with
respect to $Q$, $\KL{P}{Q}$ denotes the KL divergence. 
For the normed space $(\mathcal{X},\|\cdot\|)$, we write
$\mathcal{B}^{\mathcal{X},\|\cdot\|}(c,r):=\{x\in\mathcal{X}:\ \|x-c\|\le r\}$
for the norm ball of radius $r$ centered at $c$.

% Notation paragraph unchanged — omitted here for brevity.
\section{Setup }
\label{sec:setup}

\subsection{Model}
\begin{figure}[hbt]
    \centering
    \begin{tikzpicture}[every node/.style={font=\small}]
    \node[draw,rounded corners,align=center,text width=3.8cm,minimum height=1.5cm] (z) at (0,0)
       {Unknown label sequence\\$z^\star\in\setz(\N)\subseteq\{-1,+1\}^{\N}$};
    \node[draw,rounded corners,align=center,text width=4.0cm,minimum height=1.5cm] (p) at (5.2,0)
       {Two shared linear vectors\\$\gamma^\star_{+1},\gamma^\star_{-1}$\\One shared matrix $\Theta^\star$};
    \node[draw,rounded corners,align=center,text width=3.8cm,minimum height=1.5cm] (y) at (10.4,0)
       {Observed $Y^{(1)},\ldots,Y^{(\N)}$\\Independent given $z^\star$};
    \draw[->,thick] (z) -- (p);
    \draw[->,thick] (p) -- (y);
    \end{tikzpicture}
    \caption{\footnotesize Structured two-component model. Sample $k$ has
    linear parameter $\gamma^{\star(k)}=\gamma^\star_{z^{\star(k)}}$ and
    interaction matrix $\Theta^\star$. The estimator recovers the shared
    matrix and sample-specific parameters; recovery of both component
    vectors and labels requires the conditions in
    Corollary~\ref{cor:pooled-components}.}
    \label{fig:model-distributions}
\end{figure}
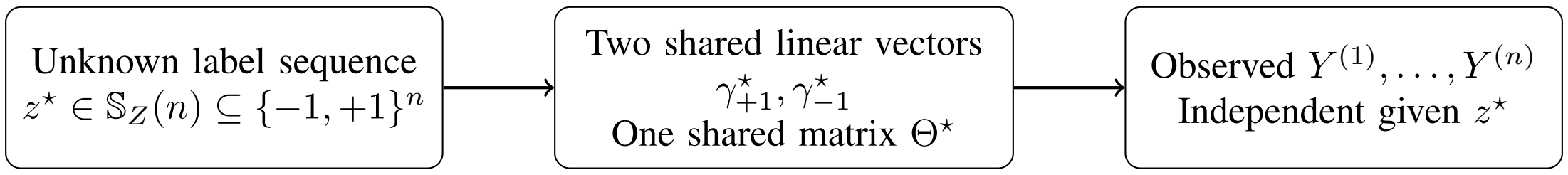

There are $\N$ samples of interest: $\{\Zk, \Yk\}_{k \in [\N]}$; refer to Figure~\ref{fig:model-distributions}. 
The $\{\Zk\}_{k \in [\N]}$ are latent or unobserved while $\{\Yk\}_{k \in [\N]}$ are observed. Next, we 
describe the model for how these samples are generated.

\renewcommand{\zprime}{+1}
\renewcommand{\zdprime}{-1}
\renewcommand{\wprime}{w}

\smallskip
\noindent \emph{Latent $\{\Zk\}_{k \in [\N]}$.} We posit that each $\Zk \in \{\zprime,\zdprime\}$
for all $k\in [\N]$ and $\{\Zk\}_{k \in [\N]} \in \setz(\N) \subseteq \{\zprime,\zdprime\}^\N$ for 
some {\em known} set $\setz(\N)$. The selection of $\{\Zk\}_{k \in [\N]} \in \setz(\N)$ is arbitrary,
however: the sequence may be fixed, or random with support in this set. The complexity of $\setz(\N)$ will impact the ability to learn. 

\smallskip
\noindent \emph{Conditional Distribution $\{\Yk\}_{k \in [\N]} | \{\Zk\}_{k \in [\N]}$.} 
Given $\{\Zk\}_{k \in [\N]} \in \setz(\N)$, the $\{\Yk\}_{k \in [\N]}$ are independent. 
For each $k \in [\N]$, $\Yk \in \mathcal{Y}^{\T} \subseteq \R^{\T}, ~\T \geq 1$ and is sampled
as per a distribution in an exponential family with linear and quadratic sufficient statistics.
The parameter for linear sufficient statistics for $\Yk$ depends on $\Zk$ while the parameter for 
quadratic sufficient statistics does not depend on $\Zk$ or does not change with $k \in [\N]$. 
With that, let $\gamma^\star = (\gamma^\star_{\zprime}, \gamma^\star_{\zdprime})$ with 
$ \gamma^\star_{\zprime}, \gamma^\star_{\zdprime} \in \mathbb{R}^{\T}$ represent possible 
linear parameters and let $\Theta^{\star} \in \R^{\T \times \T}$ denote the shared quadratic 
parameter, which is symmetric with zero diagonal. Then the density of $\Yk | \Zk$ is given as follows: 
for $y \in \mathcal{Y}^{\T}, z \in \{\zprime, \zdprime\}$, 
\begin{align}
&f_{Y|Z}\bigl( y \mid z ; \Theta^\star, \gamma^\star\bigr)
\propto
\exp\Bigl\{
    y^{\top}   \gamma^{\star}_{z}+ y^{\top} \Theta^\star y 
\Bigr\}.\label{eq:conditional-gamma-star}
\end{align}
We use $Y\mid Z=z$ as shorthand.
Given \eqref{eq:conditional-gamma-star}, it follows that for any $\es\in[\T]$, the conditional distribution of $Y_{\es}$
given $Y_{-\es} := \{Y_{\esprime}\}_{\esprime\in[\T]\setminus\{\es\}}$ and $Z$ equals 
%$\Yk_{\es}$ given $\Yk_{-\es} := \{\Yk_{\esprime}\}_{\esprime\in[\T]\setminus\{\es\}}$ and $Z^k$ equals
\begin{align}
&f_{Y_{\es} | Y_{-\es}, Z}\!\left(
     y_\es \,\middle|\,
    y_{-\es},
    z;\, \Theta^\star, \gamma^\star
\right)\propto
\exp\left\{
    \left(\gamma^\star_{z,\es} + 2 \sum_{\esprime \in [\T]\setminus\{\es\}} \Theta^\star_{\es \esprime}\, y_{\esprime}\right)
      y_\es \right\}. \label{eq:conditional-likelihood-node}
\end{align}

\smallskip
\noindent In summary, given $\{\Zk\}_{k \in [\N]}$, $\{\Yk\}_{k \in [\N]}$ are independent of each other 
conditioned on $\{\Zk\}_{k \in [\N]}$ and for each $k \in [\N]$, $\Yk | \Zk$ obeys the conditional distribution 
given by \eqref{eq:conditional-gamma-star}.

% \begin{figure}[!t]
%     \centering
%     \includegraphics[width=1.0\linewidth]
%     {isit-template/figures/example-causal.pdf}
%     \caption{\footnotesize Motivation from a causal inference perspective.}
%     \label{fig:example-causal}
% \end{figure}

%which makes no specific  assumption on the distribution .% with node-specific parameters $\Zk$.
% \begin{figure}[t]
%     \centering
%     \includegraphics[width=1.0\linewidth]{isit-template/figures/motivate-causal.pdf}
%     \caption{\footnotesize Motivation from causal inference perspective}
%     \label{fig:motivate-causal}
%     % , i.e.,  $\{\gamma^{\star}(\beta'),\gamma^{\star}(\beta'')\}$, $\{\wprime,\wdprime\}$, and $\Theta^\star$.}
%\end{figure}
%\smallskip
%\noindent \emph{Key Model Assumptions.}
{
\subsection{Representative Examples}
\label{subsec:representative-examples}
\noindent Our model is motivated by  settings in which latent heterogeneity affects local parameters while the underlying interaction structure remains shared. We describe two such motivating examples in causal inference and quantum computing. To apply our theorems, the induced linear
parameter must take two values, the observations must have the bounded
product support specified above, and the label sequence must satisfy
Assumption~\ref{assum:setz}{, which is described in Section~\ref{sec:assumptions}}. These examples motivate the model structure;
our results do not establish counterfactual identification or recovery of
a general latent confounder or noise distribution.\\

\noindent\emph{Causal inference with unobserved confounding.}
A natural connection between causal inference with unobserved confounding and mixture learning arises by viewing unobserved confounders as latent modes of heterogeneity \cite{sood2026causal}. For each unit $k\in[\N]$, let $U^{(k)}$ denote an unobserved confounder, $A^{(k)}$ the observed action, and $Y^{(k)}$ the observed outcome. Denoting the concatenated action-outcome vector by
$X^{(k)} := \big[(A^{(k)})^\top,(Y^{(k)})^\top\big]^\top$,
and modeling the conditional distribution
\begin{align}
    f_{X\mid U}(x\mid u)
    \propto
    \exp\left\{
        x^\top \gamma(u)
        + x^\top \Theta x
    \right\},
\end{align}
where
\begin{align}
    \gamma(u)
    =
    \begin{bmatrix}
        \phi^a + 2\Phi^{au}u\\
        \phi^y + 2\Phi^{yu}u
    \end{bmatrix},
    \qquad
    \Theta
    =
    \begin{bmatrix}
        \Phi^{aa} & \Phi^{ay}\\
        \Phi^{ya} & \Phi^{yy}
    \end{bmatrix}.
\end{align}
\noindent Thus, the latent type $Z^{(k)}$ in
Figure~\ref{fig:model-distributions} corresponds to the unit-level
confounder $U^{(k)}$, while the observation $Y^{(k)}$ maps to the action--outcome vector $X^{(k)}$. The confounder
perturbs the linear parameter $\gamma(U^{(k)})$, while the quadratic
interaction parameter $\Theta$ remains shared across units. Marginalizing
over $U^{(k)}$ therefore induces a mixture of exponential-family
distributions. Learning this mixture provides parameters of
$f_{Y\mid A,U}$ to evaluate counterfactual distributions \cite{shah2023counterfactualinferenceunobservedconfounding,sood2026causal}.\\%under
% alternative actions $\widetilde a$
% .\\

\noindent\emph{Quantum annealing.}
Physical computing systems provide another setting where hardware
imperfections, temperature fluctuations, and environmental noise induce heterogeneity across repeated
measurements. In quantum annealers, rapidly fluctuating noise in the residual
local fields can lead to sample-to-sample variation in the effective model leading to mixtures
\cite{PRXQuantum.3.020317}. The qubits form a hardware graph $G=(V,E)$, with local fields $h_i, i \in V$ biasing
individual qubits and pairwise couplings $J_{ij}, (i,j) \in E$ set as inputs. 
For each anneal $k\in[\N]$, let $S^{(k)}$ denote the unobserved
realization of the instantaneous local-field noise and
$\sigma^{(k)}\in\{-1,+1\}^{\T}$ the observed spin configuration.
Conditioned on $S^{(k)}=s$, the readout follows a  Gibbs
distribution
\begin{align}
    p(\sigma\mid s)
    \propto
    \exp\left\{
        \sigma^\top \gamma(s)
        + \sigma^\top \Theta \sigma
    \right\},
\end{align}
where
\begin{align}
    \gamma_i(s)
    =
    \beta_i\bigl(h_i+h_i^{\rm bias}+h_i^{\rm sd}s_i\bigr),
    \qquad
    \Theta_{ij}
    =
    \beta_{ij}J_{ij},
    \quad (i,j)\in E.
\end{align}
Here, $h_i^{\rm bias}$ captures the persistent bias in the local field,
$h_i^{\rm sd}$ its instantaneous noise magnitude, and $\beta_i,\beta_{ij}$
the effective inverse-temperature scaling of the fields and couplings. The Hamiltonian functions $H_{\rm field}(\sigma,s)=\sigma^\top\gamma(s)$ and $H_{\rm coupling}(\sigma)=\sigma^\top\Theta\sigma$ respectively correspond to the local field and coupling. Thus, the latent type $Z^{(k)}$ in Figure~\ref{fig:model-distributions} corresponds to the instantaneous
local-field noise $S^{(k)}$, while the generic observation $Y^{(k)}$ maps
to the spin configuration $\sigma^{(k)}$. The noise perturbs the linear
parameter $\gamma(S^{(k)})$, while the quadratic interaction parameter
$\Theta$ remains shared across anneals. The instantaneous
local-field noise therefore induces a mixture of Ising models rather than repeated draws from
a single fixed model. Learning this mixture provides the effective field and
coupling parameters needed to characterize the output distribution for  device
calibration \cite{PRXQuantum.3.020317}.

}
\subsection{Key Assumptions}
\label{sec:assumptions}
We make the following assumptions:
\begin{assumption}\label{assum:bounded-support} 
 {$\exists y_{\rm \max}>0$ s.t. $ \mathcal{Y} \subseteq  [-y_{\rm \max},y_{\rm \max}]$}.% and $\mathcal{Y}$ is closed.}
  % {$\exists y_{\rm \max}>0$ s.t. $\forall y \in \mathcal{Y}$, $|y|\leq y_{\rm \max}$.}
\end{assumption}
Assumption~\ref{assum:bounded-support} imposes a bounded-support exponential-family \eqref{eq:conditional-likelihood-node} structure, which is a common modeling choice when domain knowledge imposes limits on the magnitude of observations.
% \begin{assumption}
% For each $s\in[t]$, for any constants $a_0\in\R$ and $\{a_{s'}\}_{s'\in[t]\setminus\{s\}},
% \subset\R$, such that not all of them are $0$, we have
% \begin{align}
% \pr\left(\left\{a_0 + \sum_{s'\in[t]\setminus\{s\}} a_{s'} Y_{s'}
%  = 0\right\}\right) & = 0.
% \end{align}
% %holds a.s. for the measure corresponding to \eqref{eq:conditional-gamma-star} {only if} $a_0=0$ and $a_{s'}=0$  $\forall s'\in[t]\setminus\{s\}$. 
% \label{assum:identifiability}
% \end{assumption}

% \begin{assumption}
%     $w,1-w > w_{\min}$
% \end{assumption}

% \begin{figure}[t]
%     \centering
%     \includegraphics[width=1.0\linewidth]
%     {isit-template/figures/example-hardware.pdf}
%     \caption{\footnotesize Motivation from a hardware perspective.}
%     \label{fig:example-hardware}
% \end{figure}
\begin{assumption}\label{assum:identifiability}
For each $z\in\{-1,+1\}$ and any real coefficients
$a_0$, $\{a_s\}_{s\in[\T]}$, and $\{a_{ss'}\}_{1\leq s<s'\leq\T}$,
not all zero, we have
\begin{align}
\pr\Bigl(
a_0+\sum_{s\in[\T]}a_sY_s+\sum_{1\le s<s'\le\T}a_{ss'}Y_sY_{s'}=0
{\color{black}~\Big|~Z=z}
\Bigr)
{\color{black}<1}.
\label{eq:identifiability-quadratic}
\end{align}
\color{black}
\end{assumption}
Note, for each $s\in[\T]${\color{black}, each $z\in\{-1,+1\}$}, for any constants $a_0\in\mathbb R$ and
$\{a_{s'}\}_{s'\in[\T]\setminus\{s\}}\subset\mathbb R$ with $a_0^2 + \|a\|_2 > 0$,  we have
\begin{align}
\pr\left(\left\{a_0+\sum_{s'\in[\T]\setminus\{s\}}a_{s'}Y_{s'}=0\right\}{\color{black}~\bigg|~Z=z}\right){\color{black}<1}. \label{cor:identifiability-affine}
\end{align}
%\pr\Bigl(
% a_0+\langle a,\phi_{\rm ref}(Y)\rangle=0
% \Bigr)
% =
To see \eqref{cor:identifiability-affine}, apply \eqref{eq:identifiability-quadratic} with the coefficient on $Y_s$ set to $0$ and all
quadratic coefficients $a_{rr'}$ set to $0$. Since $a_0$ and $\{a_{s'}\}_{s'\ne s}$ are not all
zero by hypothesis, the resulting $(a_0,a)$ in \eqref{eq:identifiability-quadratic} is not all
zero, so the stated probability is {\color{black}less than $1$}.  Assumption~\ref{assum:identifiability} {\color{black}is a non-degeneracy condition on the linear and quadratic sufficient statistics: within each mixture component, no non-trivial linear combination of $1$, $\{Y_s\}_{s\in[\T]}$ and $\{Y_sY_{s'}\}_{1\le s<s'\le\T}$ is almost surely equal to zero. It} leads to identifiability.

% for the  distribution \eqref{eq:q-def} based on centering sufficient statistics in \eqref{eq:conditional-likelihood-node}.
%
\begin{assumption}\label{assum:delsep}
Let 
\begin{align}
     \|\gamma^{\star}_{\zprime} - \gamma^{\star}_{\zdprime}\|_{2} \geq \delsep > 0, \label{eq:gamma-sep}
\end{align}
%\begin{align}
%     \|\gamma^{\star}_{\zprime} - \gamma^{\star}_{\zdprime}\|_{2} \geq \delsep > 0~ \text{~and } \min\{ w^{\star}_{\zprime},w^{\star}_{\zdprime} \}=: w_{\min}>0. \label{eq:gamma-sep}
%\end{align}
with $\gamma^{\star}_{\zprime}, \gamma^{\star}_{\zdprime}\in \Lambda_\gamma$, where
\begin{align}
\Lambda_{\gamma}
&:=\Bigl\{\gamma \in \mathbb{R}^{\T}:\ \|\gamma\|_\infty \le \alpha\Bigr\}, \label{eq:def-lambda-gamma}
\end{align}
and $\Theta^\star \in \Lambda_\Theta$, where
\begin{align}
% &\qquad ~\exists z \in \mathcal{Z} {\rm ~s.t.~}\gamma= \sum_{d \in [\pz]} z_d \Phi^{\star}_{\rm YZ}(:,d)   \\
\Lambda_{\Theta}
&:=\Bigl\{\Theta \in \mathbb{R}^{\T\times \T}:\ 
\Theta=\Theta^\top,\|\Theta\|_{\max}\le \alpha ,\ \Theta_{\es\es}=0,\ \|\Theta_\es \|_{1}\le \beta ~\forall \es\in[\T]
\Bigr\}, \label{eq:def-lambda-Theta}
\end{align}
Here $\alpha>0$ and $\beta\geq0$ are fixed parameter bounds.
The row $\ell_1$ constraint does not impose a sparsity constraint:
it also permits dense matrices with small entries. %Further, $\Theta^\star \in \Lambda_\Theta$.
%{\color{black} talk about $\lambda_{min}>0$}
\end{assumption}

\medskip
\noindent The condition \eqref{eq:gamma-sep} ensures component density parameters are well separated. Interpreting $\Theta$ as a weighted adjacency matrix, $\|\Theta\|_{\max}\le\alpha$ bounds edge weights and the condition $\|\Theta_\es\|_1\le\beta$ as a bound on the maximum (weighted) node degree when the joint distribution of element of $Y$ is viewed as a graphical model.

\begin{assumption}\label{assum:setz}
Given $\setz(\N) \subseteq \{\zprime,\zdprime\}^{\N}$, we require that for some $\zeta\in [0,1)$,
\begin{align}\label{eq:setzsize}
    \log |\setz(\N)| & = O(\N^\zeta).
\end{align}
\end{assumption}

\medskip \noindent Assumption~\ref{assum:setz} limits the uncertainty
in the latent sequence. Nondegenerate iid Bernoulli labels have full
support $\{-1,+1\}^{\N}$ and logarithmic cardinality $\N\log2$, so they
are outside this assumption. In contrast, an unknown change point gives
only $\N+1$ assignments; see Example~\ref{ex:change-point}.  

\subsection{Goal}\label{subsec:learning-goal}
Given observations $\{y^{(k)}\}_{k\in[\N]}$ and the known class
$\setz(\N)$, our primary goal is to estimate $\Theta^\star$ and the
sample-specific parameters
$\gamma^{\star(k)}=\gamma^\star_{z^{\star(k)}}$, measured jointly by
Frobenius interaction error and average squared linear-parameter error.
When both components are observed, we also seek to estimate their two
linear vectors and labels up to permutation, under the additional
conditions stated in Corollary~\ref{cor:pooled-components}. An unobserved
component cannot be estimated. No population mixing weights are assumed
in this structured-sequence model.

\section{Main Results}
\label{sec:results}

In this section, we describe the estimation algorithm followed by the performance
guarantees.% for its ability
%to learn mixture densities and mixing proportions.

\subsection{Estimation Algorithm}

We define an M-estimator by minimizing a convex loss over the known
parameter constraints. The feasible set is generally a nonconvex union,
with one convex subproblem for each admissible label assignment. We start
by describing the loss. This is inspired by the global
interaction screening objective \cite{Vuffray-lokhov, shah2023computationally,
shah2023counterfactualinferenceunobservedconfounding}.

\smallskip
\noindent \emph{Convex Loss.}
We utilize a convex loss function building upon the prior work in
\cite{shah2023counterfactualinferenceunobservedconfounding}. Given $\{\yk\}_{k \in
[\N]}$, the loss function $\mathcal{L}_{\N,\T}: \Lambda_{\Theta} \times
\Lambda_\gamma^{[\N]} \rightarrow \mathbb{R}$ is defined as
\begin{align}
&\losssingle=
\frac{1}{\N}
\sum_{k\in[\N]} \sum_{\es\in[\T]}
\exp\Bigl\{
    -\Bigl(\gammak_\es
      + 2 \sum_{\esprime \in [\T]\setminus\{\es\}} \Theta_{\es \esprime}\, \yk_{\esprime}\Bigr) \bigl(\yk_\es - \mu_{\mathcal U}\bigr)
\Bigr\} \label{eq:mloss-single-panel}
\end{align}
where $\mu_{\mathcal U}:= \mathbb{E}_{Y\sim
\mathcal{U}_{\mathcal{Y}}}[Y]$ is the mean of the uniform reference
measure defined in the notation above. Note that the term $\bigl(\yk_\es -
\mu_{\mathcal U}\bigr)$ corresponds to a \emph{centered}
sufficient statistic that does not scale with dimension $\T$. The key computational
advantage of the convex objective \eqref{eq:mloss-single-panel} is that it bypasses
computation of the log partition function (as in Maximum Likelihood Estimation or MLE)
by introducing the centered version of the sufficient statistics and retaining
convexity. We note that each term $\exp\{
    -\bigl(\gammak_\es
+ 2 \sum_{\esprime \in [\T]\setminus\{\es\}} \Theta_{\es \esprime}\,
      \yk_{\esprime}\bigr)\bigl(\yk_\es -
      \mu_{\mathcal U}\bigr)
\}$ in \eqref{eq:mloss-single-panel} corresponds to the M-estimator
\cite{shah2023computationally} adapted to the conditional density
\eqref{eq:conditional-likelihood-node}.

\smallskip
\noindent \emph{Constraints.} We wish to estimate parameters $\Theta^\star$, 
$\{\gamma^\star = (\gamma^\star_{\zprime}, \gamma^\star_{\zdprime})\}$ and 
unobserved $\{z^{(k)}\}_{k \in [\N]}$. The shared parameter $\Theta^\star$ is
present in all $\N$ samples, while in the sample $k \in [\N]$, the linear
parameter is $\gamma^\star_{z^{(k)}}$. That is, the search space for 
parameters is 
\begin{align*}
\const(\N)
=\bigcup_{z\in\setz(\N)}
\Bigl\{\bigl(\Theta,\{\gamma_{z^{(k)}}\}_{k\in[\N]}\bigr):
\Theta\in\Lambda_\Theta,\ (\gamma_{+1},\gamma_{-1})\in\Lambda_\gamma^2\Bigr\}.
\end{align*}

\smallskip
\noindent \emph{$M$-estimation.} The parameters are estimated by minimizing loss function 
defined over the parameter space. Specifically, the estimates are 
\begin{align}
&(\widehat{\Theta},\{\widehat{\gamma}^{(k)}\}_{k\in[\N]})\in
\argmin_{\substack{\big(\Theta, \{\gamma^{(k)}\}_{k\in[\N]}\big) \in \const(\N)}}
\losssingle. \label{eq:minima-for-single-panel-loss}
\end{align}
For each fixed assignment $z\in\setz(\N)$, substitute
$\gamma^{(k)}=\gamma_{z^{(k)}}$ and minimize over the convex set
$\Lambda_\Theta\times\Lambda_\gamma^2$. The global minimum in
\eqref{eq:minima-for-single-panel-loss} is the smallest of these
$|\setz(\N)|$ minima. Thus a known assignment requires one convex
problem; a polynomial-size assignment class admits enumeration of
polynomially many convex problems. Under the model assumptions and for fixed dimension,
Assumption~\ref{assum:setz} yields statistical consistency; it does not
guarantee polynomial-time enumeration, which requires $| \setz(\N)| = O(\log \N)$.

\subsection{Guarantees}
We provide statistical guarantees on quality of estimates assuming that we have access
to an exact solution to the optimization problem \eqref{eq:minima-for-single-panel-loss}.
Recall that in our setting, given latent types $\{\Zk = \zk\}_{k\in[\N]}$, the outcomes
$\{\Yk\}_{k=1}^{\N}$ are generated independently as per
\eqref{eq:conditional-gamma-star} with
true parameters $\Theta^\star\in\Lambda_\Theta$ and $\gamma^{\star}_{\zk}\in\Lambda_\gamma$
for all $k\in[\N]$.
We provide guarantees in two steps. First, we provide guarantees in an idealized setting
with access to the population
distribution conditional on the true label sequence, corresponding to
{\em conditional population identification}. This is stated as
Theorem \ref{thm:idealized-loss}.
Second, we provide finite-sample estimation error based on the observed outcomes. This
is stated jointly for the interaction and linear parameters in
Theorem~\ref{thm:two-valued-linear-param}, with component and label guarantees
in Corollary~\ref{cor:pooled-components}.

%%%
\iffalse
%%%

\begin{table*}[!ht]
\caption{Notation summary: Estimation Algorithm}
\raggedright
\renewcommand{\arraystretch}{1.25}
\setlength{\tabcolsep}{6pt}
\begin{tabularx}{\textwidth}{
>{\raggedright\arraybackslash}p{0.40\textwidth}
X
}
\hline
\textbf{Symbol} & \textbf{Definition} \\
\hline
$\losssingle$ & Finite sample convex loss defined in \eqref{eq:mloss-single-panel} \\
$\lossidealized$ & Idealized population loss obtained as expectation of convex loss
w.r.t. the distribution $\Pi_{k \in [\N]}f_{\Yk|\Zk}\bigl(. \bigr)$,\\
$\Thetahat$ &  Estimated shared interaction parameter \\
$\widehat{I}_{\zprime},\widehat{I}_{\zdprime}$ & Estimated partition index sets \\
$\widehat{w}_{\zprime},\widehat{w}_{\zdprime}$ & Estimated mixing weights \\
$\gammakhat$ &  Estimated linear parameter for sample $k$ \\
$\widehat{\gamma}_{\zprime},\widehat{\gamma}_{\zdprime}$ & Refined estimate for linear parameters
\\
%$\widehat{\gamma}^{\rm ref}_{\zprime},\widehat{\gamma}^{\rm ref}_{\zdprime}$ & Refined
%estimate for linear parameters \\
\hline
\end{tabularx}
\label{tab:notation-method}
\end{table*}
%%
\fi
%%
%\subsubsection{Identification}

\medskip
\noindent{\bf Identification.} Consider the following \emph{idealized} version of loss function in \eqref{eq:mloss-single-panel}:
given
$\Zk = \zk \in \{\zprime, \zdprime\}$ for $k \in [\N]$, consider
\begin{align}
\lossidealized & = \mathbb{E}_{\Theta^*, \gamma^*}\Bigl[{\cal L}_{n,t}\big(\{ \Yk \}_{k \in [\N]}; \Theta, \{\gamma^{(k)}\}_{k \in [\N]}\big) ~|~ \{\zk\}_{k \in [\N]} \Bigr]
\label{eq:loss-idealized}
\end{align}
In the above, $\mathbb{E}_{\Theta^*, \gamma^*}[\cdot ~|~ \{\zk\}_{k \in [\N]}]$
represents the expectation with respect to the true underlying joint distribution of
$\{\Yk\}_{k \in [\N]}$ given $\{\Zk\}_{k \in [\N]} = \{\zk\}_{k \in [\N]}$:
for each $k \in [\N]$, $\Yk | \Zk = \zk$ is sampled as per
\eqref{eq:conditional-gamma-star} with parameters $\Theta^\star, \gamma^\star_{\zk}$
independently.

In effect, \eqref{eq:loss-idealized} can be viewed as variant of
\eqref{eq:mloss-single-panel}
where for each $k \in [n]$ with given $\Zk$, we have access to infinite number of
samples from the conditional distribution of $Y | \Zk$ and can therefore
compute its exact expectation.
We shall argue that access to this idealized loss allows for recovery of the shared matrix
$\Theta^\star$ and each observed sample's parameter $\gamma^\star_{\zk}$.

\smallskip
\begin{theorem}[Conditional population identification]
\label{thm:idealized-loss}
Let Assumptions~\ref{assum:bounded-support}--\ref{assum:delsep} hold, and
fix any label sequence $z=(z^{(k)})_{k\in[\N]}\in\{-1,+1\}^{\N}$.
The idealized loss in \eqref{eq:loss-idealized}, conditional on this
sequence, has the unique minimizer over
$\Lambda_\Theta\times\Lambda_\gamma^{[\N]}$ given by
\[
 (\widetilde\Theta,\{\widetilde\gamma^{(k)}\}_{k\in[\N]})
 =(\Theta^\star,\{\gamma^\star_{z^{(k)}}\}_{k\in[\N]}).
\]
\end{theorem}

\smallskip \noindent Theorem~\ref{thm:idealized-loss} holds on the
larger product parameter space without imposing $\const(\N)$ or
Assumption~\ref{assum:setz}. Its criterion uses the true conditional law
for every sample, so this statement does not establish identifiability
from an unrestricted marginal mixture density{;} nor does it identify the
parameter of a component absent from the label sequence. { Recall, in our problem setting}, only one observation is available for each sample. The next
theorem uses the shared parameters and admissible assignment class to
turn conditional population separation into finite-sample recovery.

\medskip
\noindent{\bf Finite-Sample, Non-Asymptotic Guarantees.} Define 
%En route to providing estimation guarantees for Algorithm~\ref{alg:two-step} using
%\cite{abhin-thesis}, we quantify 
the effective dimension of the parameter space through the notions of $\varepsilon$-nets, 
$\varepsilon$-covering numbers, and {metric entropy}. 
%defined below.

\smallskip \noindent {\em $\varepsilon$-net.} Let ${W}\subseteq \mathbb{R}^{d}$. 
A set ${U}\subseteq {W}$ is an $\varepsilon$-net of ${W}$ with respect to the
$\|\cdot\|$ norm, denoted by $\mathcal{V}_{W}(\varepsilon)$, if $\forall w\in {W},\ \exists u\in {U}
\text{ such that } \|w-u\| \leq \varepsilon.$

\smallskip \noindent {\em Covering number.} The $\varepsilon$-covering number of ${W}$, denoted by
$\covernum{W}{1,1}{\varepsilon}$, is the minimum cardinality over all $\varepsilon$-nets of ${W}$.

\smallskip \noindent {\em Metric entropy.} For a set ${W}$ with $\varepsilon$-covering number
$\covernum{W}{1,1}{\varepsilon}$, its metric entropy with respect to the $\|\cdot\|$ norm is defined as
$
\mentropy{W}{1,1}{\varepsilon}
=
\log\bigl(\covernum{W}{1,1}{\varepsilon}\bigr).
$

\smallskip
% Linear-parameter recovery exploits the two shared component vectors.
We next exploit the fact that, for each fixed assignment in $\setz(\N)$,
only two linear parameter vectors need to be estimated. The resulting guarantee
controls the shared interaction matrix and the sample-averaged linear
parameter error jointly. All probabilities below are conditional on an
arbitrary true assignment $z^\star=(z^{\star(k)})_{k\in[\N]}\in\setz(\N)$;
therefore the same guarantees hold unconditionally when the true assignment
belongs to this set almost surely. Let $\mu_{\mathcal U}=\E_{\mathcal U_{\mathcal Y}}[Y]$, $\bar Y_s=Y_s-\mu_{\mathcal U}$,
and define
\begin{align}
\widetilde Y_s&:=\bar Y_s(1,(Y_{s'})_{s'\neq s})^\top,\qquad
\lambda_*:=\min_{\substack{u\in\{-1,+1\}\\s\in[\T]}}
\lambda_{\min}\!\left(\E[\widetilde Y_s\widetilde Y_s^\top\mid Z=u]\right)>0,
\label{eq:pooled-feature-curvature}\\
M_\tau&:=\exp\{2y_{\max}(\alpha+2\beta y_{\max})\},\qquad
 a_*:=\frac{\lambda_*}{2M_\tau},\qquad
 b_*:=M_\tau\max\{2y_{\max},4y_{\max}^2\}.
\label{eq:pooled-constants}
\end{align}
Positivity of $\lambda_*$ follows from Assumption~\ref{assum:identifiability}.
 Here $\mathcal M_{\Lambda_\Theta}$ uses the entrywise $\ell_1$ norm
$\|\Theta\|_{1,1}=\sum_{s,s'}|\Theta_{ss'}|$, and
$\mathcal M_{\Lambda_\gamma}$ uses the vector $\ell_1$ norm. For a candidate $p=(\Theta,\{\gamma^{(k)}\}_{k\in[\N]})$ and the true tuple
$p^\star=(\Theta^\star,\{\gamma^{\star(k)}\}_{k\in[\N]})$, where
$\gamma^{\star(k)}=\gamma^\star_{z^{\star(k)}}$, measure the joint estimation
error by
\begin{align}
 d_\N(p,p^\star)^2
 :=4\|\Theta-\Theta^\star\|_{\mathrm F}^2
   +\frac1\N\sum_{k\in[\N]}\|\gamma^{(k)}-\gamma^{\star(k)}\|_2^2.
\label{eq:pooled-parameter-distance}
\end{align}
Thus $d_\N$ controls the interaction matrix in Frobenius norm and the
linear parameters in root mean squared error across samples.

\begin{theorem}[Joint recovery with two shared linear parameters]
\label{thm:two-valued-linear-param}
Let Assumptions~\ref{assum:bounded-support}--\ref{assum:setz} hold, with
$z^\star\in\setz(\N)$, and let $\widehat p$ be any global minimizer of
\eqref{eq:minima-for-single-panel-loss}. Fix a target accuracy
$\varepsilon>0$ and a failure probability $\delta\in(0,1)$, and set
$\rho_\varepsilon=a_*\varepsilon^2/(12b_*)$.
If
\begin{align}
\N\geq\frac{16\T M_\tau^3}{a_*\varepsilon^2}
\left\{\log\frac{|\setz(\N)|}{\delta}
+\mathcal M_{\Lambda_\Theta}(\rho_\varepsilon)
+2\mathcal M_{\Lambda_\gamma}(\rho_\varepsilon)\right\},
\label{eq:two-valued-linear-param-sample-complexity}
\end{align}
then, with probability at least $1-\delta$, every such minimizer satisfies
\begin{align}
 d_\N(\widehat p,p^\star)^2
 =4\|\widehat\Theta-\Theta^\star\|_{\mathrm F}^2
 +\frac1\N\sum_{k\in[\N]}
 \|\widehat\gamma^{(k)}-\gamma^\star_{z^{\star(k)}}\|_2^2
 \leq\varepsilon^2.
\label{eq:two-valued-linear-param-recovery}
\end{align}
In particular, the interaction error satisfies
\begin{align}
 \|\widehat\Theta-\Theta^\star\|_{2,\infty}
 :=\max_{s\in[\T]}\|(\widehat\Theta-\Theta^\star)_s\|_2
 \leq\|\widehat\Theta-\Theta^\star\|_{\mathrm F}
 \leq\varepsilon/2,
\label{eq:pooled-interaction-recovery}
\end{align}
and the root mean squared linear-parameter error is at most $\varepsilon$.
\end{theorem}

\Needspace{10\baselineskip}
\begin{example}[An unknown change point]
\label{ex:change-point}
Suppose the observations are ordered and, for an unknown
$r \in\{0,1,\ldots,\N\}$, the true labels satisfy
\[
 z_{k}^{(r)}=
 \begin{cases}+1,&1\leq k\leq r ,\\-1,&r<k\leq\N.\end{cases}
\]
Then
\[
 \setz(\N)=\{z^{(r)}:0\leq r\leq\N\},\qquad
 |\setz(\N)|=\N+1,\qquad \log|\setz(\N)|=\log(\N+1).
\]
This class satisfies Assumption~\ref{assum:setz} for every fixed
$\zeta\in(0,1)$; retaining the exact logarithm gives a sharper bound than
replacing it by $O(\N^\zeta)$.

\smallskip
\noindent\emph{Computation.}
For each $r\in\{0,\ldots,\N\}$, minimize
\begin{align}
 \mathcal L^{(r)}(\Theta,\gamma_+,\gamma_-)
 :=\frac1\N\left\{
 \sum_{k=1}^{r}\sum_{s=1}^{\T}\ell_s(Y^{(k)};\Theta,\gamma_+)
 +\sum_{k=r+1}^{\N}\sum_{s=1}^{\T}\ell_s(Y^{(k)};\Theta,\gamma_-)
 \right\}
\label{eq:changepoint-objective}
\end{align}
over $\Lambda_\Theta\times\Lambda_\gamma^2$, where $\ell_s$ is the
coordinate summand in \eqref{eq:mloss-single-panel}.
For a fixed $r$ this is a convex optimization problem. Comparing its
minimum over all $\N+1$ values of $r$ yields the global estimator in
\eqref{eq:minima-for-single-panel-loss}, with the minimizing $r$ providing
an estimate of the change point. This requires only $\N+1$ convex
subproblems.

\smallskip
\noindent\emph{An explicit sample-size bound.}
Let $t \geq 2$ and $d_\Theta=\T(\T-1)/2$. The parameter constraints imply
$\|\Theta\|_{1,1}\leq\T\beta$ and $\|\gamma\|_1\leq\T\alpha$.
A packing argument in the respective normed spaces gives
\begin{align}
 \mathcal M_{\Lambda_\Theta}(\rho)
 &\leq d_\Theta\log\left(1+\frac{2\T\beta}{\rho}\right),\qquad
 \mathcal M_{\Lambda_\gamma}(\rho)
 \leq\T\log\left(1+\frac{2\T\alpha}{\rho}\right).
\label{eq:changepoint-entropy}
\end{align}
For any certified lower bound $0<a\leq a_*$, define
\begin{align}
 \rho&:=\frac{a\varepsilon^2}{12b_*},\qquad
 A:=\max\left\{1,\frac{16\T M_\tau^3}{a\varepsilon^2}\right\},
 \nonumber\\
 L&:=\log\frac1\delta
   +\frac{\T(\T-1)}2\log\left(1+\frac{24b_*\T\beta}{a\varepsilon^2}\right)
   +2\T\log\left(1+\frac{24b_*\T\alpha}{a\varepsilon^2}\right).
\label{eq:changepoint-constants}
\end{align}
Then the fully explicit sufficient condition
\begin{align}
 %\boxed
 {\quad\N\geq
 \left\lceil2A\bigl(L+\log(2A)\bigr)\right\rceil\quad}
\label{eq:changepoint-explicit-n}
\end{align}
ensures, with probability at least $1-\delta$,
\begin{align}
 4\|\widehat\Theta-\Theta^\star\|_{\mathrm F}^2
 +\frac1\N\sum_{k=1}^{\N}
    \|\widehat\gamma^{(k)}-\gamma^\star_{z^{\star(k)}}\|_2^2
 \leq\varepsilon^2.
\label{eq:changepoint-guarantee}
\end{align}
There is no unknown $\N$ on the right-hand side of
\eqref{eq:changepoint-explicit-n}. To verify it, substitution in
Theorem~\ref{thm:two-valued-linear-param} reduces the requirement to
$\N\geq A\{L+\log(\N+1)\}$, which is sufficient even when the maximum
in the definition of $A$ selects $1$. Since $A\geq1$, the inequality
$\log x\leq x-1$, applied to $x=(\N+1)/(2A)$, gives
\[
 \log(\N+1)\leq\frac{\N}{2A}+\log(2A).
\]
Condition~\eqref{eq:changepoint-explicit-n} therefore implies the required
sample-size bound. A smaller curvature lower bound $a\leq a_*$ only
increases the prefactor and decreases the covering radius, so it remains
sufficient for the theorem.

\smallskip
\begin{samepage}
\noindent\emph{Dependence on dimension.}
Let $a=a_*$. If
$\lambda_*\geq\lambda_0>0$ uniformly in $\T$, and
$\alpha,\beta,y_{\max},\lambda_0$ are fixed, the sufficient sample size
in \eqref{eq:changepoint-explicit-n} has order
\begin{align}
 \frac{\T^3}{\varepsilon^2}
       \log\left(2+\frac{\T}{\varepsilon^2}\right)
 +\frac{\T}{\varepsilon^2}\log\frac1\delta,
 \qquad 0<\varepsilon\leq1.
\label{eq:changepoint-rate-conditioned}
\end{align}
\end{samepage}
\noindent A dimension-independent curvature bound is not required. To give a
bound depending only on $\T,\varepsilon,\delta$ and the fixed support
and parameter bounds, let
\begin{align}
 v_{\mathcal U}&:=\operatorname{Var}_{\mathcal U_{\mathcal Y}}(Y)>0,
 \qquad \eta:=v_{\mathcal U}/M_\tau,\nonumber\\
 \underline\lambda_{\T}
 &:=\frac{\eta^2}
 {2\eta+(\T-1)\{1+2(\T-1)y_{\max}^2\}},\qquad
 a_{\T}:=\frac{\underline\lambda_{\T}}{2M_\tau}.
\label{eq:changepoint-explicit-curvature}
\end{align}
Section~\ref{sec:change-point-details} proves
$\lambda_*\geq\underline\lambda_{\T}$.
Thus using $a=a_{\T}$ in \eqref{eq:changepoint-constants} gives an explicit
bound without an unknown feature eigenvalue. For fixed
$\alpha,\beta,y_{\max},v_{\mathcal U}>0$, its sufficient sample size has order
\begin{align}
 \frac{\T^5}{\varepsilon^2}
       \log\left(2+\frac{\T}{\varepsilon^2}\right)
 +\frac{\T^3}{\varepsilon^2}\log\frac1\delta,
 \qquad 0<\varepsilon\leq1.
\label{eq:changepoint-rate-uniform}
\end{align}
\end{example}

\smallskip
\noindent\emph{Implications of Theorem \ref{thm:two-valued-linear-param}.}
The metric entropy in \eqref{eq:two-valued-linear-param-sample-complexity} 
accounts for uncertainty in one interaction matrix and two linear vectors. 
Searching over the admissible assignments adds $\log|\setz(\N)|$. To make 
the resulting requirement on $\N$ explicit, write Assumption~\ref{assum:setz} as
$\log|\setz(\N)|\leq C_Z\N^\zeta$ for $\N\geq n_0$, where
$C_Z>0$ and $0\leq\zeta<1$. Define
\[
 A_\varepsilon:=\frac{16\T M_\tau^3}{a_*\varepsilon^2},\qquad
 D_{\varepsilon,\delta}:=\log\frac1\delta
 +\mathcal M_{\Lambda_\Theta}(\rho_\varepsilon)
 +2\mathcal M_{\Lambda_\gamma}(\rho_\varepsilon).
\]
A sufficient condition for the theorem is
\begin{align}
 \N\geq\max\left\{n_0,
       (2A_\varepsilon C_Z)^{1/(1-\zeta)},
       2A_\varepsilon D_{\varepsilon,\delta}\right\}.
\label{eq:pooled-explicit-sample-size}
\end{align}
Indeed, this is satisfied for $n$ large enough given fixed $\varepsilon> 0, \delta \in (0,1)$ and $t \geq 1$. 

\smallskip
\noindent\emph{General error bound and known assignments.}
The proof in Section~\ref{sec:proof-pooled-mixture} also gives the following
form, which is useful for stating component and label guarantees.
For any fixed $\rho>0$, define
\begin{align}
H_\N(\rho,\delta)
&:=\log\frac{|\setz(\N)|}{\delta}
 +\mathcal M_{\Lambda_\Theta}(\rho)
 +2\mathcal M_{\Lambda_\gamma}(\rho),
\label{eq:pooled-entropy}\\
B_\N(\rho,\delta)
&:=\frac{1}{a_*}\left\{\frac{8\T M_\tau^3}{\N}
 H_\N(\rho,\delta)+6b_*\rho\right\}.
\label{eq:pooled-error-bound}
\end{align}
With probability at least $1-\delta$,
\begin{align}
 d_\N(\widehat p,p^\star)^2\leq B_\N(\rho,\delta).
\label{eq:pooled-general-recovery}
\end{align}
The choice $\rho=\rho_\varepsilon$ and the sample-size condition in the
theorem give $B_\N(\rho_\varepsilon,\delta)\leq\varepsilon^2$.
If the assignment is known, so that $\setz(\N)=\{z^\star\}$, write
$n_u=|\{k:z^{\star(k)}=u\}|$. The same event gives
\begin{align}
\sum_{u\in\{-1,+1\}}\frac{n_u}{\N}
\|\widehat\gamma_u-\gamma_u^\star\|_2^2&\leq B_\N(\rho,\delta),
\label{eq:pooled-known-labels}\\
\|\widehat\gamma_u-\gamma_u^\star\|_2
&\leq\sqrt{\frac{\N B_\N(\rho,\delta)}{n_u}}
\quad(n_u>0).
\nonumber
\end{align}
In particular, under the theorem's sample-size condition, this component
error is at most $\varepsilon\sqrt{\N/n_u}$.
For fixed $\T$ and fixed true parameters, the compact finite-dimensional
parameter sets have entropy $O(\log(1/\rho))$ as $\rho\downarrow0$.
Taking $\rho=1/\N$ and using Assumption~\ref{assum:setz} gives
\begin{align}
B_\N(1/\N,\delta)
=O\!\left(\N^{\zeta-1}+\frac{\log\N+\log(1/\delta)}{\N}\right).
\label{eq:pooled-consistency-rate}
\end{align}
The constants may depend on $\T$, the parameter bounds, and $\lambda_*$.
Thus the average squared error tends to zero. When the labels are known,
the assignment-entropy term vanishes and each occupied component benefits
from pooling its samples.

For unknown labels, an average error guarantee does not by itself imply
that every sample is correctly assigned. The next consequence states the
component and label guarantees precisely. Define
\[
 w_{\min,\N}:=\min_{u\in\{-1,+1\}}\frac{n_u}{\N},\qquad
 d_Z(z^\star):=
 \min_{\substack{z\in\setz(\N)\\z\notin\{z^\star,-z^\star\}}}
 \min_{\sigma\in\{-1,+1\}}
 |\{k:\sigma z^{(k)}\neq z^{\star(k)}\}|,
\]
with the minimum over an empty set equal to $+\infty$.

\begin{corollary}[Component estimates and label recovery]
\label{cor:pooled-components}
Under Theorem~\ref{thm:two-valued-linear-param}, suppose both true components
are observed, i.e. $w_{\min,\N} > 0$, and let $B=B_\N(\rho,\delta)$. If
$B<w_{\min,\N}\delsep^2/4$, then, on the event in
\eqref{eq:pooled-general-recovery}, any representation
$\widehat\gamma^{(k)}=\widehat\gamma_{\widehat z^{(k)}}$ with
$\widehat z\in\setz(\N)$ admits a permutation $\pi$ of $\{-1,+1\}$ such that
\begin{align}
\|\widehat\gamma_{\pi(u)}-\gamma_u^\star\|_2
&\leq\sqrt{\frac{\N B}{n_u}},\qquad u\in\{-1,+1\},
\label{eq:pooled-component-recovery}\\
\frac1\N|\{k:\widehat z^{(k)}\neq\pi(z^{\star(k)})\}|
&\leq\frac{4B}{\delsep^2}.
\label{eq:pooled-label-error}
\end{align}
If additionally $4\N B/\delsep^2<d_Z(z^\star)$, all labels are recovered
up to $\pi$, and
\begin{align}
\max_{k\in[\N]}\|\widehat\gamma^{(k)}-\gamma_{z^{\star(k)}}^\star\|_2
\leq\sqrt{B/w_{\min,\N}}.
\label{eq:pooled-uniform-recovery}
\end{align}
\end{corollary}

\section{Conclusions}\label{sec:conclusions}

We developed a method for learning mixture of two bounded-support exponential-family
components with a shared interaction matrix and two linear parameter
vectors. The estimator is an M-estimator that utilizes a computationally tractable 
convex loss function. The conditional population loss identifies the interaction matrix
and the parameter attached to each sample. For a global empirical minimizer, 
our finite-sample guarantee controls Frobenius interaction error
and average squared linear-parameter error, with an assignment-complexity
cost of $\log|\setz(\N)|$. For fixed dimension, sublinear growth of this
cost yields consistency. Component and label recovery additionally depend
on component occupancy, parameter separation, and, for exact labels,
separation between admissible partitions. %Future 
{Future work includes extending the analysis to more than two components and higher-order polynomial sufficient statistics. Another direction is to extend estimation to settings with overlapping mixture components. Deploying and extending the framework to settings with different forms of latent heterogeneity holds promise, including accounting for unobserved confounding in causal inference and calibrating physical and quantum computing systems with fluctuating local fields.}

\newpage
\bibliographystyle{IEEEtran}
\bibliography{ref-merged}

\section*{Appendix}
%\section{Proof Sketch for Theorem~\ref{thm:idealized-loss}}
\section{Proof of Theorem~\ref{thm:idealized-loss}}
\label{sec:thm1-sketch-brief}

Fix a label sequence $z=(z^{(k)})_{k\in[\N]}$. Conditional on this
sequence, the observations are independent and $Y^{(k)}$ has density
\eqref{eq:conditional-gamma-star} with parameters
$\Theta^\star,\gamma^\star_{z^{(k)}}$. The empirical loss depends only on
the observed samples and candidate parameters. Its idealized counterpart
is a theoretical conditional expectation under the true law, which is
not itself available from the observations.
Recalling from \eqref{eq:loss-idealized}, we have
\begin{align*}
\lossidealized & = \mathbb{E}_{\Theta^*, \gamma^*}\Bigl[{\cal L}_{n,t}\big(\{ \Yk \}_{k \in [\N]}; \Theta, \{\gamma^{(k)}\}_{k \in [\N]}\big) ~|~ \{\zk\}_{k \in [\N]} \Bigr].
\end{align*}
Define 
\begin{align}
\ell_{\es}\big(y; \Theta, \gamma\big) & =\exp\Biggl\{
    -\Bigl(\gamma_{\es}
      + 2 \sum_{\esprime \in [\T]\setminus\{\es\}} \Theta_{\es \esprime}\, y_{\esprime}\Bigr)
      \bigl(y_{\es} - \mu_{\mathcal U}\bigr)
\Biggr\}.\label{eq:loss-idealized-kt}
\end{align}
Then, we can re-write 
\begin{align*}
    {\cal L}_{n,t}\big(\{ \Yk \}_{k \in [\N]}; \Theta, \{\gamma^{(k)}\}_{k \in [\N]}\big)
    & = \frac{1}{n}\sum_{k \in [\N]}\sum_{s \in [\T]} \ell_{\es}\big(\Yk; \Theta, \gammak\big).
\end{align*}
Therefore, 
\begin{align}\label{eq:thm.1.1}
\lossidealized & = \frac1n \sum_{k \in [\N]} \sum_{s \in [t]} \mathbb{E}_{\Theta^*, \gamma^*}\Bigl[ \ell_{\es}\big(\Yk; \Theta, \gammak\big) ~|\zk \Bigr].
\end{align}
For each $\es \in [\T]$, consider a ``modified'' density over $\mathcal{Y}^{\T}$ given $z \in \{\zprime, \zdprime\}$ with parameters $\Theta \in \mathbb{R}^{\T \times \T}$, $\gamma \in \mathbb{R}^{\T}$ as 
\begin{align}\label{eq:thm.1.2}
    q_{\es}(y | z; \Theta, \gamma) & = \frac{ f_{Y | Z}(y | z; \Theta^*, \gamma^\star) \ell_{\es}\big(y; \Theta, \gamma\big)}{
    \mathbb{E}_{\Theta^*, \gamma^\star}[\ell_{\es}\big(\cdot; \Theta, \gamma\big) | z]}, 
\end{align}
where recall from \eqref{eq:conditional-gamma-star} that 
\begin{align}
    f_{Y | Z}(y | z; \Theta^*, \gamma^\star) & \propto \exp(y^T \gamma^\star_z + y^T \Theta^\star y ),
\end{align}
and $\mathbb{E}_{\Theta^*, \gamma^\star}[\cdot | z]$ is an expectation with respect to this distribution 
over $\mathcal{Y}^{\T}$. With these notations and definitions, we state two key Lemmas whose proofs are provided subsequently. 
\begin{lemma}\label{lemma:dkl}
For each $\es \in [\T]$, given $\Theta \in \mathbb{R}^{\T \times \T}, \gamma \in \mathbb{R}^{\T}$ and $z \in \{\zprime, \zdprime\}$,
\begin{align}
\mathbb{E}_{\Theta^*, \gamma^\star}[\ell_{\es}\big(\cdot; \Theta, \gamma\big) | z]
& = g_{\es}(z; \Theta^*, \gamma^*) 
\exp \Big( 
\KL{q_{\es}(\cdot | z; \Theta^\star, \gamma^\star_z)}{q_{\es}(\cdot | z; \Theta, \gamma)}
\Big)
\label{eq:lemma-kl}
\end{align}
where $g_{\es}(z; \Theta^*, \gamma^*) > 0$ and does not depend on $\Theta, \gamma$. 
\end{lemma}
\begin{lemma}\label{lem:diff}
For each $\es \in [\T]$, given $\Theta \in \mathbb{R}^{\T \times \T}, \gamma \in \mathbb{R}^{\T}$ and $z \in \{\zprime, \zdprime\}$,
\begin{align}
q_{\es}(y | z; \Theta, \gamma) & = q_{\es}(y | z; \Theta^\star, \gamma^\star_z), \quad\text{for almost every }y\in\mathcal Y^{\T}
\end{align}
if and only if 
{\color{black}
$\gamma_{\es} = \gamma^\star_{z,\es}$ and
$\Theta_{\es\esprime}=\Theta^\star_{\es\esprime}$
for all $\esprime \in [\T]\setminus\{\es\}$.
}
\end{lemma}

Given Lemmas~\ref{lemma:dkl} and~\ref{lem:diff}, Theorem~\ref{thm:idealized-loss} can be established as follows. Note that the parameters and outcomes are both bounded (Assumptions~\ref{assum:bounded-support} and \ref{assum:delsep}), and the density \eqref{eq:thm.1.2} is strictly positive on $\mathcal{Y}^{\T}$.
Recall that $\KL{.}{.}\ge 0$, with equality if and only if its arguments
are equal almost everywhere with respect to the reference measure. Therefore, by Lemmas~\ref{lemma:dkl}
and \ref{lem:diff} as well as \eqref{eq:thm.1.1}, {\color{black}applying Lemma~\ref{lem:diff} for each $\es\in[\T]$,} it follows that $\tilde{\cal L}$ attains its minimum if and only if 
$\Theta = \Theta^\star$ and $\gamma^{(k)} = \gamma^\star_{\zk}$ for all $k \in [\N]$. This completes the 
proof of Theorem \ref{thm:idealized-loss}.
\myendpf

\subsection{Proof of Lemma~\ref{lemma:dkl}}
Consider a fixed $\es \in [\T]$. For any given $\Theta \in \mathbb{R}^{\T \times \T}, \gamma \in \mathbb{R}^{\T}$
and $z \in \{\zprime, \zdprime\}$, we shall establish that

\begin{align}
&\log\left(\mathbb{E}_{\Theta^*, \gamma^\star}[\ell_{\es}\big(\cdot; \Theta, \gamma\big) | z]\right)\nonumber\\
&\qquad \qquad =  
\KL{q_{\es}(\cdot | z; \Theta^\star, \gamma^\star_z)}{q_{\es}(\cdot | z; \Theta, \gamma)} 
- 
\KL{q_{\es}(\cdot | z; \Theta^\star, \gamma^\star_z)}{f_{Y | Z}(\cdot | z; \Theta^*, \gamma^\star)}.
%\qPsinoK{\cdot}{\Psi_{\es}^\star}}{\qPsinoK{\cdot}{\Psi_{\es}}}-\KL{\qPsinoK{\cdot}{\Psi_{\es}^\star}}{\fYkZkPsinoK{\cdot}{\Psi^{\star}}} 
\label{eq:equiv-claim1}
\end{align}
The \eqref{eq:equiv-claim1} will establish Lemma \ref{lemma:dkl} with 
\begin{align}
    g_{\es}(z; \Theta^\star, \gamma^\star) & = 
    \exp\Big(-\KL{q_{\es}(\cdot | z; \Theta^\star, \gamma^\star_z)}{f_{Y | Z}(\cdot | z; \Theta^*, \gamma^\star)}\Big).
\end{align}
Therefore, to conclude the proof of Lemma \ref{lemma:dkl}, we need to establish \eqref{eq:equiv-claim1}. Towards that, consider
\begin{align}
&\KL{q_{\es}(\cdot | z; \Theta^\star, \gamma^\star_z)}{q_{\es}(\cdot | z; \Theta, \gamma)} 
- 
\KL{q_{\es}(\cdot | z; \Theta^\star, \gamma^\star_z)}{f_{Y | Z}(\cdot | z; \Theta^*, \gamma^\star)}\nonumber
\\
&\qquad =
\int_{\mathcal{Y}^\T}
q_{\es}(y | z; \Theta^\star, \gamma^\star_z)
\log\!\left(
\frac{q_{\es}(y | z; \Theta^\star, \gamma^\star_z)}{q_{\es}(y | z; \Theta, \gamma)}
\frac{f_{Y | Z}(y | z; \Theta^*, \gamma^\star)}{q_{\es}(y | z; \Theta^\star, \gamma^\star_z)}
\right)\,dy \nonumber \\
&\qquad =
\int_{\mathcal{Y}^\T}
q_{\es}(y | z; \Theta^\star, \gamma^\star_z)
\log\!\left(
\frac{f_{Y | Z}(y | z; \Theta^*, \gamma^\star)}{q_{\es}(y | z; \Theta, \gamma)} \right)\,dy. \label{eq:kl-interim1}
\end{align}
From \eqref{eq:thm.1.2}, we have 
\begin{align}
\frac{ f_{Y | Z}(y | z; \Theta^*, \gamma^\star) }{ q_{\es}(y | z; \Theta, \gamma)}
& = \frac{\mathbb{E}_{\Theta^*, \gamma^\star}[\ell_{\es}\big(\cdot; \Theta, \gamma\big) | z]}{\ell_{\es}\big(y; \Theta, \gamma\big)}.
\label{eq:kl-interim2}
\end{align}
Using \eqref{eq:kl-interim2} in \eqref{eq:kl-interim1} together with the definition of the loss \eqref{eq:loss-idealized-kt}, we have 
\begin{align}
&\KL{q_{\es}(\cdot | z; \Theta^\star, \gamma^\star_z)}{q_{\es}(\cdot | z; \Theta, \gamma)} 
- 
\KL{q_{\es}(\cdot | z; \Theta^\star, \gamma^\star_z)}{f_{Y | Z}(\cdot | z; \Theta^*, \gamma^\star)}\nonumber
\\
&\qquad =
\log\left( \mathbb{E}_{\Theta^*, \gamma^\star}[\ell_{\es}\big(\cdot; \Theta, \gamma\big) | z] \right)
+ \int_{\mathcal{Y}^\T} 
q_{\es}(y | z; \Theta^\star, \gamma^\star_z) \big(- \log \ell_{\es}\big(y; \Theta, \gamma\big)\big)\, dy \nonumber \\
&\qquad =
\log\left( \mathbb{E}_{\Theta^*, \gamma^\star}[\ell_{\es}\big(\cdot; \Theta, \gamma\big) | z] \right)
+ \int_{\mathcal{Y}^\T} 
q_{\es}(y | z; \Theta^\star, \gamma^\star_z) \Bigl(\gamma_{\es}
      + 2 \sum_{\esprime \in [\T]\setminus\{\es\}} \Theta_{\es \esprime}\, y_{\esprime}\Bigr) \bigl(y_{\es} - \mu_{\mathcal U}\bigr)\, dy. \label{eq:lemma1.0}
\end{align}
Therefore, to conclude the proof of Lemma \ref{lemma:dkl}, we would like to conclude that \eqref{eq:lemma1.0} equals $$\log\left( \mathbb{E}_{\Theta^*, \gamma^\star}[\ell_{\es}\big(\cdot; \Theta, \gamma\big) | z] \right),$$ i.e. the second term equals $0$. 
To do so, we will need two properties. First, notice that 
\begin{align}
    \int_{\mathcal{Y}} \bigl(y_{\es} - \mu_{\mathcal U}\bigr) dy_{\es} & = 0. \label{eq:zero-mean}
\end{align}
Next, the term $\Bigl(\gamma_{\es}
      + 2 \sum_{\esprime \in [\T]\setminus\{\es\}} \Theta_{\es \esprime}\, y_{\esprime}\Bigr)$ does not contain $y_{\es}$. 
Therefore, if we argue that $q_{\es}(y | z; \Theta^\star, \gamma^\star_z)$ does not depend on $y_{\es}$, then 
\begin{align}
& \int_{\mathcal{Y}^\T} q_{\es}(y | z; \Theta^\star, \gamma^\star_z) \Bigl(\gamma_{\es}
      + 2 \sum_{\esprime \in [\T]\setminus\{\es\}} \Theta_{\es \esprime}\, y_{\esprime}\Bigr) \bigl(y_{\es} - \mu_{\mathcal U}\bigr)\, dy \nonumber \\
    & \qquad = 
    \Bigl(\int_{\mathcal{Y}^{\T-1}} q_{\es}(y | z; \Theta^\star, \gamma^\star_z) \Bigl(\gamma_{\es}
      + 2 \sum_{\esprime \in [\T]\setminus\{\es\}} \Theta_{\es \esprime}\, y_{\esprime}\Bigr) dy_{-s}\Bigr) ~\times~
    \Bigl(\int_{\mathcal{Y}} \bigl(y_{\es} - \mu_{\mathcal U}\bigr) dy_{\es}\Bigr)\nonumber \\
    & \qquad = 0.
\end{align}
To establish that $q_{\es}(y | z; \Theta^\star, \gamma^\star_z)$ does not depend on $y_{\es}$, from \eqref{eq:thm.1.2} we 
have
\begin{align}
    q_{\es}(y | z; \Theta^\star, \gamma^\star_z) 
    & \propto \ell_{\es}\big(y; \Theta^\star, \gamma^\star_z\big) f_{Y | Z}(y | z; \Theta^*, \gamma^\star)  \nonumber \\
    & \propto \exp\Biggl\{-\Bigl(\gamma^\star_{z,\es} + 2 \sum_{r'\in[\T]\setminus\{\es\}}\Theta^\star_{\es r'}\,y_{r'}\Bigr) \bigl(y_{\es}-\mu_{\mathcal U}\bigr) 
    + \sum_{r\in[\T]}\gamma^\star_{z,r}\,y_r +2 \sum_{1\le r<r'\le \T}\Theta^\star_{rr'}\,y_ry_{r'} \Biggr\} \nonumber\\
   & = \exp\Biggl\{\Bigl(\gamma^\star_{z,\es} + 2 \sum_{r'\in[\T]\setminus\{\es\}}\Theta^\star_{\es r'}\,y_{r'}\Bigr) \mu_{\mathcal U} 
    + \sum_{r\in[\T]\setminus \{\es\}} \gamma^\star_{z,r}\,y_r +2 \sum_{1\le r<r'\le \T: r,r'\neq \es }\Theta^\star_{rr'}\,y_ry_{r'} \Biggr\}.
\end{align}
This concludes the proof of Lemma \ref{lemma:dkl}.

\subsection{Proof of Lemma~\ref{lem:diff}}
Given $z \in \{\zprime, \zdprime\}$, $\Theta \in \mathbb{R}^{\T \times \T}, \gamma \in \mathbb{R}^{\T}$,
from \eqref{eq:thm.1.2}, it follows that
\begin{align}
\frac{q_{\es}(y | z; \Theta^\star, \gamma^\star_z)}
{q_{\es}(y | z; \Theta, \gamma)}
& \propto \frac{\ell_{\es}\big(y; \Theta^\star, \gamma^\star_z\big)}
{\ell_{\es}\big(y; \Theta, \gamma\big)}\nonumber \\
& = \exp\Biggl\{
{\color{black}
-\Bigl(
\bigl(\gamma^{\star}_{z, \es}-\gamma_{\es}\bigr)
+ 2 \sum_{\esprime \in [\T]\setminus\{\es\}} y_{\esprime}
  \bigl(\Theta^\star_{\es \esprime}-\Theta_{\es \esprime}\bigr)
\Bigr)
}
\cdot\bigl(y_{\es}
- \mu_{\mathcal U}\bigr)
\Biggr\},
\label{eq:likelihood-ratio}
\end{align}
Therefore, {\color{black}if}
$q_{\es}(y | z; \Theta^\star, \gamma^\star_z)
= q_{\es}(y | z; \Theta, \gamma)$ for
almost every $y \in \mathcal{Y}^{\T}$, {\color{black}then} it must be that the ratio on the right hand side of
\eqref{eq:likelihood-ratio}
is constant almost everywhere. Consider the function
\begin{align}
\nu(y) &=
\Bigl(
\bigl(\gamma^{\star}_{z, \es}-\gamma_{\es}\bigr)
{\color{black}
+ 2 \sum_{\esprime \in [\T]\setminus\{\es\}} y_{\esprime}
  \bigl(\Theta^\star_{\es \esprime}-\Theta_{\es \esprime}\bigr)
}
\Bigr)
\cdot\bigl(y_{\es}
- \mu_{\mathcal U}\bigr).
\end{align}

{\color{black}
\noindent There must  exist a constant $c \in \mathbb{R}$ such that
$\nu(y)-c=0$ for almost every $y\in\mathcal Y^{\T}$.
Expanding $\nu(y)-c$, we have
\begin{align}
0
&=
\bigl(\gamma^{\star}_{z,\es}-\gamma_{\es}\bigr)y_{\es}
-
\bigl(\gamma^{\star}_{z,\es}-\gamma_{\es}\bigr)
\mu_{\mathcal U}
+2\sum_{\esprime\in[\T]\setminus\{\es\}}
\bigl(\Theta^\star_{\es\esprime}-\Theta_{\es\esprime}\bigr)
y_{\es}y_{\esprime}
\nonumber\\
&\quad
-2\mu_{\mathcal U}
\sum_{\esprime\in[\T]\setminus\{\es\}}
\bigl(\Theta^\star_{\es\esprime}-\Theta_{\es\esprime}\bigr)
y_{\esprime}
-c. \label{eq:interim}
\end{align}
{Since \eqref{eq:interim} holds almost everywhere under the reference measure, it holds almost surely under $Y\mid Z=z$, i.e., the probability in \eqref{eq:identifiability-quadratic} equals $1$ for the corresponding coefficient vector. By} Assumption~\ref{assum:identifiability}, we have {that coefficient vector must be zero; and noting the coefficients of $y_\es y_{\esprime}$ and of $y_\es$ gives} that
$\Theta^\star_{\es\esprime}=\Theta_{\es\esprime}$ for all
$\esprime\neq\es$ and
$\gamma^\star_{z,\es}=\gamma_{\es}$.

\noindent Conversely, if
$\Theta^\star_{\es\esprime}=\Theta_{\es\esprime}$ for all
$\esprime\neq\es$ and
$\gamma^\star_{z,\es}=\gamma_{\es}$, then
\begin{align}
\ell_{\es}\big(y;\Theta^\star,\gamma^\star_z\big)
=
\ell_{\es}\big(y;\Theta,\gamma\big),
\qquad \forall y\in\mathcal{Y}^{\T},
\end{align}
and hence, from \eqref{eq:thm.1.2},
\begin{align}
q_{\es}(y | z; \Theta^\star, \gamma^\star_z)
=
q_{\es}(y | z; \Theta, \gamma),
\qquad \forall y\in\mathcal{Y}^{\T}.
\end{align}
Therefore,
$q_{\es}(y | z; \Theta^\star, \gamma^\star_z)
= q_{\es}(y | z; \Theta, \gamma)$ for almost every
$y \in \mathcal{Y}^{\T}$ if and only if
$\Theta^\star_{\es\esprime}=\Theta_{\es\esprime}$ for all
$\esprime\neq\es$ and
$\gamma^\star_{z,\es}=\gamma_{\es}$.
}This completes the proof of Lemma~\ref{lem:diff}.
\section{Proof of Theorem~\ref{thm:two-valued-linear-param}}
\label{sec:proof-pooled-mixture}

%\subsection*{Step 1: identify the true parameter and the comparison to prove}
Fix the true assignment $z^\star=(z^{\star(k)})_{k\in[\N]}\in\setz(\N)$.
The true sample-specific linear parameters are
\[
 \gamma^{\star(k)}:=
 \begin{cases}
  \gamma_+^\star,&z^{\star(k)}=+1,\\
  \gamma_-^\star,&z^{\star(k)}=-1,
 \end{cases}
 \qquad
 p^\star:=\left(\Theta^\star,
                 \{\gamma^{\star(k)}\}_{k\in[\N]}\right).
\]
Although $z^\star$, $\Theta^\star$, and the two component vectors are
unknown to the estimator, they are fixed in this analysis. Throughout the
proof, probabilities and expectations are conditional on $z^\star$.
Conditional on this assignment, the observations are independent: the
$n_+$ observations with label $+1$ share $\gamma_+^\star$, and the $n_-$
observations with label $-1$ share $\gamma_-^\star$. Both groups share
$\Theta^\star$. Consider a candidate tuple 
\[
 p:=\left(\Theta,\{\gamma^{(k)}\}_{k\in[\N]}\right).
\]
Let $\mathcal L_{\N,\T}(p)$ denote the empirical loss
\eqref{eq:mloss-single-panel}, with the observations suppressed from the
notation, and define the empirical and population loss differences
\begin{align}
 \widehat m(p)&:=\mathcal L_{\N,\T}(p)-\mathcal L_{\N,\T}(p^\star),
 \nonumber\\
 m(p)&:=\E[\mathcal L_{\N,\T}(p)\mid z^\star]
       -\E[\mathcal L_{\N,\T}(p^\star)\mid z^\star].
\label{eq:pooled-excess-definitions}
\end{align}
By Theorem \ref{thm:idealized-loss}, the true tuple $p^\star$ minimizes the population loss. 
It need not minimize the empirical loss for a particular observed data set. Our task is to show
that a candidate far from $p^\star$ cannot minimize that empirical loss.

More precisely, we will prove that, on a single event of probability at
least $1-\delta$, every admissible $p\in\const(\N)$ satisfies
\begin{align}
 \widehat m(p)\geq\frac{a_*}{2}
       \left(d_\N(p,p^\star)^2-\varepsilon^2\right)
\label{eq:pooled-target-gap}
\end{align}
under conditions of this theorem. Thus $d_\N(p,p^\star)>\varepsilon$ implies
$\mathcal L_{\N,\T}(p)>\mathcal L_{\N,\T}(p^\star)$.

\medskip 
\noindent{\em Bounding empirical loss by parameter error for a fixed parameter.}
Fix a deterministic candidate $p$ in the ambient parameter space
$\Lambda_\Theta\times\Lambda_\gamma^{[\N]}$. It is not chosen as a function
of the observations. This step does not yet use the two-component
restriction; we will use that restriction when taking a union bound over feasible
candidates. 

\smallskip
\noindent\emph{(a) The population loss gap controls the parameter error.}
Write the parameter errors as
\[
 \Omega:=\Theta-\Theta^\star,\qquad
 \omega^{(k)}:=\gamma^{(k)}-\gamma^{\star(k)}.
\]
For sample $k$ and coordinate $s$, introduce the true conditional field
and its change under the candidate:
\begin{align*}
 h_{s,*}^{(k)}&:=\gamma_s^{\star(k)}
           +2\sum_{s'\neq s}\Theta_{ss'}^\star Y_{s'}^{(k)},\\
 D_s^{(k)}&:=\omega_s^{(k)}
           +2\sum_{s'\neq s}\Omega_{ss'}Y_{s'}^{(k)},
 \qquad V_s^{(k)}:=D_s^{(k)}\bar Y_s^{(k)}.
\end{align*}
The quantity $D_s^{(k)}$ is the error in the conditional field, and
$V_s^{(k)}$ is the corresponding error in the exponent of the loss.
Recall
\begin{align}
 \ell_s(Y;\Theta_s,\gamma_s)
 :=\exp\left\{-\left(\gamma_s+
           2\sum_{s'\neq s}\Theta_{ss'}Y_{s'}\right)
           (Y_s-\mu_{\mathcal U})\right\}.
\label{eq:pooled-coordinate-loss}
\end{align}
Set
\[
 L_k(p):=\sum_{s\in[\T]}\ell_s(Y^{(k)};\Theta_s,\gamma_s^{(k)}),\qquad
 X_k(p):=L_k(p)-L_k(p^\star).
\]
By \eqref{eq:mloss-single-panel},
$\mathcal L_{\N,\T}(p)=\N^{-1}\sum_kL_k(p)$, so
$\widehat m(p)=\N^{-1}\sum_k X_k(p)$.

To compare $L_k(p)$ and $L_k(p^\star)$, interpolate between their
parameters. At interpolation value $v\in[0,1]$, the parameters are
$\Theta^\star+v\Omega$ and
$\gamma^{\star(k)}+v\omega^{(k)}$. The resulting loss is
\[
 F_k(v):=\sum_s
 \exp\{-(h_{s,*}^{(k)}+vD_s^{(k)})\bar Y_s^{(k)}\}.
\]
Thus $F_k(0)=L_k(p^\star)$, $F_k(1)=L_k(p)$, and
$X_k(p)=F_k(1)-F_k(0)$. The segment lies in the convex ambient parameter
space\footnote{It need not stay in $\const(\N)$, which is not needed for this
calculation.}.  By Assumptions \ref{assum:bounded-support}-\ref{assum:delsep}
\[
 |h_{s,*}^{(k)}+vD_s^{(k)}|\leq\alpha+2\beta y_{\max},
 \qquad |\bar Y_s^{(k)}|\leq2y_{\max}.
\]
Consequently every exponential on the segment is between
$M_\tau^{-1}$ and $M_\tau$. Differentiating gives
\begin{align*}
 F_k'(0)&=-\sum_s V_s^{(k)}e^{-h_{s,*}^{(k)}\bar Y_s^{(k)}},\\
 F_k''(v)&=\sum_s(V_s^{(k)})^2
              e^{-(h_{s,*}^{(k)}+vD_s^{(k)})\bar Y_s^{(k)}}
           \geq M_\tau^{-1}\sum_s(V_s^{(k)})^2.
\end{align*}

We next verify that the first derivative has mean zero at the truth.
Conditional on $Y_{-s}^{(k)}$ and $z^\star$, both $D_s^{(k)}$ and
$h_{s,*}^{(k)}$ are fixed. The true conditional law of $Y_s^{(k)}$
has density proportional to $e^{h_{s,*}^{(k)}y}$ with respect to
$\mathcal U_{\mathcal Y}$. Hence, writing $h=h_{s,*}^{(k)}$,
\begin{align*}
 \E\!\left[\bar Y_s^{(k)}e^{-h\bar Y_s^{(k)}}
             \mid Y_{-s}^{(k)},z^\star\right]%\\
 &=
 \frac{e^{h\mu_{\mathcal U}}}
      {\int e^{hy}\,\mathcal U_{\mathcal Y}(dy)}
 \int(y-\mu_{\mathcal U})\,\mathcal U_{\mathcal Y}(dy)=0.
\end{align*}
Multiplying by $D_s^{(k)}$, summing over $s$, and removing the
conditioning on $Y_{-s}^{(k)}$ proves
$\E[F_k'(0)\mid z^\star]=0$.
Taylor's formula now yields
\begin{align*}
 X_k(p)&=F_k'(0)+\int_0^1(1-v)F_k''(v)\,dv,\\
 m_k(p):=\E[X_k(p)\mid z^\star]
 &\geq\frac1{2M_\tau}\sum_s\E[(V_s^{(k)})^2\mid z^\star]\geq0.
\end{align*}
That is
\begin{align}
 \sum_s\E[(V_s^{(k)})^2\mid z^\star]\leq2M_\tau m_k(p).
\label{eq:pooled-population-gap}
\end{align}
Now, we connect this to $d_\N(p, p^\star)$. Towards that, note that 
for any nonzero coefficient vector $c=(c_0,(c_{s'})_{s'\neq s})$,
\[
 c^\top\widetilde Y_s
 =(Y_s-\mu_{\mathcal U})
       \left(c_0+\sum_{s'\neq s}c_{s'}Y_{s'}\right)
\]
is a nonzero polynomial in the constant, linear, and pairwise quadratic
sufficient statistics. If $c_0\neq0$ its coefficient on $Y_s$ is nonzero;
otherwise some coefficient on $Y_sY_{s'}$ is nonzero.
Assumption~\ref{assum:identifiability} implies that this polynomial is not
almost surely zero in either component. Thus its squared expectation is
positive, each feature moment matrix is positive definite, and the
minimum of their finitely many smallest eigenvalues is positive.
Since $\Omega_{ss}=0$, we have
\[
 V_s^{(k)}=(\omega_s^{(k)},(2\Omega_{ss'})_{s'\neq s})^\top
                 \widetilde Y_s^{(k)},
\]
and therefore
\[
 \E[(V_s^{(k)})^2\mid z^\star]
 \geq\lambda_*\left((\omega_s^{(k)})^2
                    +4\|\Omega_s\|_2^2\right).
\]
Summing over $s$, averaging over $k$, using
$a_*=\lambda_* /(2M_\tau)$ and \eqref{eq:pooled-population-gap}, we have
\begin{align}
 m(p)\geq a_*\left(4\|\Omega\|_{\mathrm F}^2
           +\frac1\N\sum_k\|\omega^{(k)}\|_2^2\right)
       =a_*d_\N(p,p^\star)^2.
\label{eq:pooled-joint-curvature}
\end{align}

\smallskip
\noindent\emph{(b) The empirical loss retains the population gap.}
We must control how far $\widehat m(p)$ can fall below $m(p)$.
For each coordinate, the mean value theorem applied along the same
segment bounds its loss difference by $M_\tau|V_s^{(k)}|$.
Consequently, by Cauchy--Schwarz,
\[
 |X_k(p)|\leq M_\tau\sum_s|V_s^{(k)}|,\qquad
 X_k(p)^2\leq\T M_\tau^2\sum_s(V_s^{(k)})^2.
\]
Taking expectations and using \eqref{eq:pooled-population-gap} gives
\begin{align}
 \sum_k\operatorname{Var}(X_k(p)\mid z^\star)
 \leq\sum_k\E[X_k(p)^2\mid z^\star]
 \leq2\N\T M_\tau^3m(p).
\label{eq:pooled-bernstein-variance}
\end{align}
Thus the variance of the loss difference is controlled by the population
loss gap itself. This sharper variance estimate is useful when that gap
is small.

Each $L_k$ lies in $[\T/M_\tau,\T M_\tau]$, so
$|X_k(p)|\leq\T(M_\tau-M_\tau^{-1})\leq\T M_\tau$ and
$|X_k(p)-m_k(p)|\leq2\T M_\tau$.
Apply Bernstein's inequality to the independent centered variables
$X_k(p)-m_k(p)$, using \eqref{eq:pooled-bernstein-variance}.
For any $x>0$, except on an event of probability at most $e^{-x}$,
\begin{align}
 \widehat m(p)
 &\geq m(p)-\sqrt{\frac{4\T M_\tau^3m(p)x}{\N}}
              -\frac{4\T M_\tau x}{3\N}\nonumber\\
 &\geq\frac12m(p)-\frac{4\T M_\tau^3x}{\N}.
\label{eq:pooled-relative-concentration}
\end{align}
The last line follows from $2\sqrt{mA}\leq m/2+2A$ with
$A=\T M_\tau^3x/\N$, and from $M_\tau\geq1$.
Together with \eqref{eq:pooled-joint-curvature}, this gives 
\begin{align}
 \mathcal L_{\N,\T}(p)-\mathcal L_{\N,\T}(p^\star)
 \geq\frac{a_*}{2}d_\N(p,p^\star)^2
          -\frac{4\T M_\tau^3x}{\N}.
\label{eq:pooled-fixed-candidate-gap}
\end{align}
For this fixed $p$, the right-hand side is positive whenever its squared
distance from $p^\star$ exceeds $8\T M_\tau^3x/(a_*\N)$.

\medskip
\noindent{\em Extending bound on empirical loss by parameter error for all feasible parameters.} 
We extend the above bound for all feasible parameters in $\const(\N)$ next. 

\noindent\emph{(a) For a fixed assignment, cover only three shared parameters.}
For each $z\in\setz(\N)$, let
\[
 \const_z:=\left\{
  \left(\Theta,\{\gamma_{z^{(k)}}\}_{k\in[\N]}\right):
  \Theta\in\Lambda_\Theta,\ \gamma_+,\gamma_-\in\Lambda_\gamma
 \right\}.
\]
Then $\const(\N)=\bigcup_{z\in\setz(\N)}\const_z$.
Once $z$ is fixed, selecting a point in $\const_z$ means selecting only
$\Theta$, $\gamma_+$, and $\gamma_-$. Every sample with candidate label
$+1$ uses the same vector $\gamma_+$, and every sample with candidate
label $-1$ uses the same vector $\gamma_-$. Thus we need two covers of
$\Lambda_\gamma$, regardless of how large $\N$ is.

Fix a covering radius $\rho>0$. Choose a minimal $\rho$-net
$\mathcal V_\Theta\subseteq\Lambda_\Theta$ in entrywise $\ell_1$ and a
minimal $\rho$-net $\mathcal V_\gamma\subseteq\Lambda_\gamma$ in vector
$\ell_1$. For a given $z$, form
\[
 \mathcal V_z:=\left\{
  \left(\Theta_0,\{(\gamma_0)_{z^{(k)}}\}_{k\in[\N]}\right):
  \Theta_0\in\mathcal V_\Theta,\
  (\gamma_0)_+,(\gamma_0)_-\in\mathcal V_\gamma
 \right\}.
\]
For any $p\in\const_z$, choose the approximating matrix and the two
approximating vectors from these nets. Their induced tuple $p_0\in\mathcal V_z$
satisfies
\begin{align}
 \|\Theta-\Theta_0\|_{1,1}&\leq\rho,\nonumber\\
 \frac1\N\sum_k\|\gamma_{z^{(k)}}-(\gamma_0)_{z^{(k)}}\|_1
 &=\sum_{u\in\{-1,+1\}}\frac{|\{k:z^{(k)}=u\}|}{\N}
                \|\gamma_u-(\gamma_0)_u\|_1\leq\rho.
\label{eq:pooled-net-approximation}
\end{align}
The weights in the last sum add to one. This explains why the covering
radius for a shared vector does not shrink by a factor of $\N$.
The union $\mathcal V:=\bigcup_{z\in\setz(\N)}\mathcal V_z$ contains at most $K$ points, where
\begin{align}
 K:=|\setz(\N)|\,|\mathcal V_\Theta|\,|\mathcal V_\gamma|^2
 =|\setz(\N)|\exp\{\mathcal M_{\Lambda_\Theta}(\rho)
                    +2\mathcal M_{\Lambda_\gamma}(\rho)\}.
\label{eq:pooled-union-cardinality}
\end{align}

\smallskip
\noindent\emph{(b) Union bound over the covers of the admissible assignments.}
For each $p_0\in\mathcal V$, apply
\eqref{eq:pooled-relative-concentration} with
\[
 x=\log(K/\delta)=H_\N(\rho,\delta).
\]
The failure probability at one net point is at most $\delta/K$.
A union bound therefore gives an event $\mathcal E$ with probability at
least $1-\delta$, on which
\begin{align}
 \widehat m(p_0)\geq\frac12m(p_0)
              -\frac{4\T M_\tau^3}{\N}H_\N(\rho,\delta)
 \qquad\text{for every }p_0\in\mathcal V.
\label{eq:pooled-net-event}
\end{align}

\smallskip
\noindent\emph{(c) Extend from the finite cover to every feasible tuple.}
Fix any $p\in\const(\N)$, let corresponding label assignment be $z\in \setz(\N)$.
Choose $p_0\in\mathcal V_z$ as in \eqref{eq:pooled-net-approximation}.
We now prove the deterministic Lipschitz bound needed to transfer the
loss comparison from $p_0$ to $p$. The segment joining these two tuples
lies in the convex ambient parameter space. As in the fixed-candidate calculation above, each coordinate
loss on this segment is bounded above by $M_\tau$. Applying the mean
value theorem to the exponential in \eqref{eq:pooled-coordinate-loss}
therefore gives, for every sample $k$ and coordinate $s$,
\begin{align*}
 &\left|\ell_s(Y^{(k)};\Theta_s,\gamma_s^{(k)})
          -\ell_s(Y^{(k)};(\Theta_0)_s,\gamma_{0,s}^{(k)})\right|\\
 &\quad\leq M_\tau|\bar Y_s^{(k)}|
       \left(|\gamma_s^{(k)}-\gamma_{0,s}^{(k)}|
       +2\sum_{s'\neq s}|\Theta_{ss'}-(\Theta_0)_{ss'}|
                         |Y_{s'}^{(k)}|\right)\\
 &\quad\leq2y_{\max}M_\tau|\gamma_s^{(k)}-\gamma_{0,s}^{(k)}|
       +4y_{\max}^2M_\tau\|\Theta_s-(\Theta_0)_s\|_1.
\end{align*}
Summing over $s$ and averaging over $k$ yields
\begin{align*}
 |\mathcal L_{\N,\T}(p)-\mathcal L_{\N,\T}(p_0)|
 &\leq4y_{\max}^2M_\tau\|\Theta-\Theta_0\|_{1,1}
       +\frac{2y_{\max}M_\tau}{\N}
           \sum_k\|\gamma^{(k)}-\gamma_0^{(k)}\|_1.
\end{align*}
Using $b_*=M_\tau\max\{2y_{\max},4y_{\max}^2\}$ and the two bounds in
\eqref{eq:pooled-net-approximation}, we obtain
\begin{align}
 |\mathcal L_{\N,\T}(p)-\mathcal L_{\N,\T}(p_0)|
 \leq b_*\left(\|\Theta-\Theta_0\|_{1,1}
             +\frac1\N\sum_k\|\gamma^{(k)}-\gamma_0^{(k)}\|_1\right)
 \leq2b_*\rho.
\label{eq:pooled-lipschitz}
\end{align}
Taking expectations gives the same bound for the population losses.
Since the subtracted true loss is identical for both candidates,
\[
 |\widehat m(p)-\widehat m(p_0)|\leq2b_*\rho,
 \qquad |m(p)-m(p_0)|\leq2b_*\rho.
\]
On $\mathcal E$, these inequalities and \eqref{eq:pooled-net-event} imply
\begin{align}
 \widehat m(p)
 &\geq\widehat m(p_0)-2b_*\rho\nonumber\\
 &\geq\frac12m(p_0)
       -\frac{4\T M_\tau^3}{\N}H_\N(\rho,\delta)-2b_*\rho\nonumber\\
 &\geq\frac12m(p)
       -\frac{4\T M_\tau^3}{\N}H_\N(\rho,\delta)-3b_*\rho\nonumber\\
 &\geq\frac{a_*}{2}d_\N(p,p^\star)^2
       -\frac{4\T M_\tau^3}{\N}H_\N(\rho,\delta)-3b_*\rho.
\label{eq:pooled-uniform-loss-gap}
\end{align}
The event $\mathcal E$ was fixed before choosing $p$, so this bound holds
simultaneously for every $p\in\const(\N)$, including candidates selected
using the observations.

\medskip
\noindent{\em Concluding.} 
Take $\rho=\rho_\varepsilon=a_*\varepsilon^2/(12b_*)$.
Then $3b_*\rho_\varepsilon=a_*\varepsilon^2/4$.
The sample-size condition
\eqref{eq:two-valued-linear-param-sample-complexity} also gives
\[
 \frac{4\T M_\tau^3}{\N}H_\N(\rho_\varepsilon,\delta)
 \leq\frac{a_*\varepsilon^2}{4}.
\]
Substituting these two bounds into \eqref{eq:pooled-uniform-loss-gap}
proves \eqref{eq:pooled-target-gap}. Thus every feasible candidate with
$d_\N(p,p^\star)>\varepsilon$ has strictly larger empirical loss than
$p^\star$. On the other hand, feasibility of $p^\star$ and optimality
of any global minimizer $\widehat p$ imply
\[
 \mathcal L_{\N,\T}(\widehat p)
 \leq\mathcal L_{\N,\T}(p^\star),\qquad
 \widehat m(\widehat p)\leq0.
\]
Consequently $d_\N(\widehat p,p^\star)\leq\varepsilon$ on $\mathcal E$,
which proves the theorem with probability at least $1-\delta$.
The argument holds for every fixed true assignment in $\setz(\N)$ with
the same probability bound, so averaging over a random true assignment
supported on this set gives the unconditional statement as well.

\begin{samepage}
For the general bound \eqref{eq:pooled-general-recovery}, retain any fixed
$\rho>0$ and substitute $\widehat m(\widehat p)\leq0$ directly in
\eqref{eq:pooled-uniform-loss-gap}. Rearranging gives
\[
 d_\N(\widehat p,p^\star)^2
 \leq\frac1{a_*}\left\{\frac{8\T M_\tau^3}{\N}H_\N(\rho,\delta)
                       +6b_*\rho\right\}=B_\N(\rho,\delta).
\]
When $z^\star$ is known, each fitted vector is constant on its true group,
so the linear part of this distance is exactly
$\sum_u(n_u/\N)\|\widehat\gamma_u-\gamma_u^\star\|_2^2$.
This proves \eqref{eq:pooled-known-labels} and its componentwise consequence.
\hfill$\square$
\end{samepage}

\medskip
\noindent\emph{Proof of Corollary~\ref{cor:pooled-components}.}
Given \eqref{eq:pooled-general-recovery}, we have for $u \in \{\zprime, \zdprime\}$, 
\[
 \min_{v\in\{-1,+1\}}\|\widehat\gamma_v-\gamma_u^\star\|_2^2
 \leq\frac1{n_u}\sum_{k:z^{\star(k)}=u}
       \|\widehat\gamma_{\widehat z^{(k)}}-\gamma_u^\star\|_2^2
 \leq\frac{\N B}{n_u}.
\]
Under $B<w_{\min,\N}\delsep^2/4$, each true vector is within strictly less
than $\delsep/2$ of a fitted vector. The two selected fitted vectors must
be distinct, since otherwise the triangle inequality would contradict
$\|\gamma_+^\star-\gamma_-^\star\|_2\geq\delsep$. They define the
permutation $\pi$ and prove \eqref{eq:pooled-component-recovery}.
If $z^{\star(k)}=u$ but $\widehat z^{(k)}=\pi(-u)$, then
\[
 \|\widehat\gamma^{(k)}-\gamma_u^\star\|_2
 \geq\|\gamma_{-u}^\star-\gamma_u^\star\|_2
       -\|\widehat\gamma_{\pi(-u)}-\gamma_{-u}^\star\|_2
 >\delsep/2.
\]
Summing these squared errors proves \eqref{eq:pooled-label-error}.
If $4\N B/\delsep^2<d_Z(z^\star)$, an estimated assignment outside
$\{z^\star,-z^\star\}$ would contradict the definition of $d_Z(z^\star)$.
The correct assignments and \eqref{eq:pooled-component-recovery}
give \eqref{eq:pooled-uniform-recovery}.
\hfill$\square$

\section{Example~\ref{ex:change-point}: Details}
\label{sec:change-point-details}

\noindent\emph{Covering numbers.}
In a $d$-dimensional normed space, let $W$ lie in a ball of radius $R$.
A maximal $\rho$-separated subset of $W$ is a $\rho$-net with feasible
centers. Its disjoint open $\rho/2$-balls lie in a ball of radius
$R+\rho/2$, so volume comparison gives at most $(1+2R/\rho)^d$ points.
For $\Lambda_\gamma$, take $d=\T$, $R=\T\alpha$, and the vector $\ell_1$
norm. For $\Lambda_\Theta$, the symmetric zero-diagonal subspace has
$d=\T(\T-1)/2$; use $R=\T\beta$ and the entrywise $\ell_1$ norm.
This proves \eqref{eq:changepoint-entropy}; if $R=0$ or $d=0$, the
covering number is one.

\medskip
\noindent\emph{A uniform lower bound on $\lambda_*$.}
Fix a component $u$ and suppress the conditioning on $Z=u$ throughout
this argument. Every one-coordinate conditional density relative to
$\mathcal U_{\mathcal Y}$ is $e^{hy}/\int e^{hx}\,\mathcal U_{\mathcal Y}(dx)$,
where $|h|\leq\alpha+2\beta y_{\max}$.
It is at least $e^{-2y_{\max}|h|}\geq M_\tau^{-1}$.
Therefore, for every $j$ and every constant $c$,
\[
 \E[(Y_j-c)^2\mid Y_{-j}]
 \geq M_\tau^{-1}\E_{\mathcal U}[(Y-c)^2]
 \geq\eta.
\]
In particular,
$\operatorname{Var}(Y_j\mid Y_{-j})\geq\eta$ and
$\E[(Y_j-\mu_{\mathcal U})^2\mid Y_{-j}]\geq\eta$.

Fix $s$, put $d=\T-1$, and first suppose $d\geq1$.
The feature vector is
$\widetilde Y_s=\bar Y_s(1,(Y_i)_{i\neq s})^\top$.
Thus a coefficient vector for this feature consists of a scalar $c$ and
$d$ coefficients $b=(b_i)_{i\in[\T]\setminus\{s\}}$:
\[
 (c,b)^\top\widetilde Y_s
 =\bar Y_s\left(c+\sum_{i\neq s}b_iY_i\right),\qquad
 \|b\|_2^2=\sum_{i\neq s}b_i^2.
\]
Here the entries of $b$ retain the original coordinate labels $i\neq s$;
there is no entry $b_s$. Equivalently, if $b$ is embedded in
$\mathbb R^{\T}$, its $s$-th entry is set to zero. The coefficient of
$\bar Y_s$ itself is $c$, which is treated separately below.
Therefore none of the norm of $b$ can lie in coordinate $s$.
If $b=0$, the variance bound below is immediate. Otherwise choose an
index $j\neq s$ at which $b_i^2$ is largest. The largest of these $d$
nonnegative numbers is at least their average, so
\[
 b_j^2=\max_{i\neq s}b_i^2
 \geq\frac1d\sum_{i\neq s}b_i^2
 =\frac{\|b\|_2^2}{d}.
\]

To obtain a variance bound, write $W=Y_{-\{s,j\}}$ for all coordinates
except $s$ and $j$. The conditional variance formula gives
\begin{align*}
 \operatorname{Var}(Y_j\mid W)
 &=\E[\operatorname{Var}(Y_j\mid W,Y_s)\mid W]
   +\operatorname{Var}(\E[Y_j\mid W,Y_s]\mid W)\\
 &\geq\eta,
\end{align*}
because $(W,Y_s)=Y_{-j}$ and the first conditional variance is at least
$\eta$. Given $W$, every term of
$b^\top Y_{-s}=\sum_{i\neq s}b_iY_i$ except $b_jY_j$ is fixed.
Applying the variance formula once more, now removing the conditioning
on $W$, yields
\[
\begin{aligned}
 \operatorname{Var}(b^\top Y_{-s})
 &\geq\E[\operatorname{Var}(b^\top Y_{-s}\mid W)]\\
 &=b_j^2\E[\operatorname{Var}(Y_j\mid W)]
 \geq\eta b_j^2
 \geq\frac{\eta}{d}\|b\|_2^2.
\end{aligned}
\]
This argument does not require independence among the coordinates of $Y$.

Let $m=\E[Y_{-s}]$, $c\in\mathbb R$, and
$Q=\E[(c+b^\top Y_{-s})^2]$. Then
\[
 Q\geq(c+b^\top m)^2+\frac{\eta}{d}\|b\|_2^2.
\]
To relate $Q$ to the full coefficient norm $c^2+\|b\|_2^2$, first
note that each of the $d$ coordinates of $m=\E[Y_{-s}]$ has absolute
value at most $y_{\max}$. Hence $\|m\|_2^2\leq d y_{\max}^2$.
Writing $c=(c+b^\top m)-b^\top m$ and using
$(x-y)^2\leq2x^2+2y^2$, followed by Cauchy--Schwarz, gives
\[
\begin{aligned}
 c^2+\|b\|_2^2
 &\leq2(c+b^\top m)^2+2(b^\top m)^2+\|b\|_2^2\\
 &\leq2(c+b^\top m)^2+(1+2\|m\|_2^2)\|b\|_2^2\\
 &\leq2(c+b^\top m)^2+(1+2d y_{\max}^2)\|b\|_2^2.
\end{aligned}
\]
The last step is where the bound on $\|m\|_2^2$ is used.
Next, the preceding lower bound on $Q$ implies separately that
\[
 (c+b^\top m)^2\leq Q,\qquad
 \|b\|_2^2\leq\frac d\eta Q.
\]
Substituting these two inequalities yields
\[
 c^2+\|b\|_2^2
 \leq\left(2+\frac{d(1+2d y_{\max}^2)}{\eta}\right)Q.
\]
Finally, conditioning on $Y_{-s}$ yields
\begin{align*}
 \E\!\left[\bar Y_s^2(c+b^\top Y_{-s})^2\right]
 &\geq\eta Q\\
 &\geq\frac{\eta^2}{2\eta+d(1+2d y_{\max}^2)}
                    (c^2+\|b\|_2^2).
\end{align*}
For $\T=1$, the feature consists only of $\bar Y_s$, whose second
moment is at least $\eta\geq\underline\lambda_1=\eta/2$.
The bounds hold for every coordinate and both components, proving
$\lambda_*\geq\underline\lambda_{\T}$ in all dimensions.
For fixed support and parameter bounds,
$a_{\T}^{-1}=O(\T^2)$; substitution in
\eqref{eq:changepoint-constants}--\eqref{eq:changepoint-explicit-n}
gives \eqref{eq:changepoint-rate-uniform}.
\hfill$\square$

\end{document}